\documentclass[10pt,twocolumn,letterpaper]{article}
\pdfoutput=1
\usepackage{subcaption}
\usepackage{iccv}
\usepackage{times}
\usepackage{epsfig}
\usepackage{graphicx}
\usepackage{adjustbox}

\usepackage{amsmath}
\usepackage{amssymb}

\newcommand{\tabincell}[2]{\begin{tabular}{@{}#1@{}}#2\end{tabular}}

\newcounter{savecntr}
\newcounter{restorecntr}

\iccvfinalcopy % *** Uncomment this line for the final submission

\def\iccvPaperID{5987} % *** Enter the ICCV Paper ID here

\begin{document}

%%%%%%%%% TITLE
\title{DAWN: Dual Augmented Memory Network for\\ Unsupervised Video Object Tracking}

\author{
Zhenmei Shi\setcounter{savecntr}{\value{footnote}}\thanks{Equal contribution}       \space\space\space\space\space\space\space\space Haoyang Fang\setcounter{restorecntr}{\value{footnote}}\setcounter{footnote}{\value{savecntr}}\footnotemark \setcounter{footnote}{\value{restorecntr}}\\
HKUST\\
{\tt\small \{zshiad,hfangac\}@connect.ust.hk}
\and
Yu-Wing Tai\\
Tencent Youtu\\
{\tt\small 
yuwingtai@tencent.com}
\and
Chi-Keung Tang\\
HKUST\\
{\tt\small 
cktang@cs.ust.hk}
}

\maketitle
%\thispagestyle{empty}

%%%%%%%%% ABSTRACT
\begin{abstract}
Psychological studies have found that human visual tracking system
involves learning, memory, and planning. Despite recent successes, not
many works have focused on memory and planning in deep learning based
tracking. We are thus interested in memory augmented network, where an
external memory remembers the evolving appearance of the target
(foreground) object without backpropagation for updating weights. Our Dual
Augmented Memory Network (DAWN) is unique in remembering both target and
background, and using an improved attention LSTM memory to guide the focus
on memorized features. DAWN is effective in unsupervised  tracking in
handling total occlusion, severe motion blur, abrupt changes in target
appearance, multiple object instances, and similar foreground and
background features. We present extensive quantitative and qualitative
experimental comparison with state-of-the-art methods including top
contenders in recent VOT challenges. Notably,
despite the straightforward implementation, DAWN is ranked third in both
VOT2016 and VOT2017 challenges with excellent success rate among all VOT
fast trackers running at fps $>$ 10 in unsupervised tracking in both
challenges. We propose DAWN-RPN, where we simply augment
our memory and attention LSTM modules to the state-of-the-art
SiamRPN, and report immediate performance gain, thus
demonstrating DAWN can work well with and directly benefit other models to
handle difficult cases as well.
\end{abstract}
%%%%%%%%% BODY TEXT
\section{Introduction}

\begin{figure}[thb]
    \centering
    \begin{subfigure}[b]{0.98\linewidth}
        \centering
        \includegraphics[width=0.9\linewidth]{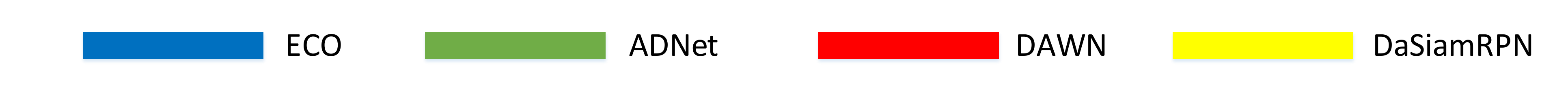}
    \end{subfigure}
    \begin{subfigure}{0.96\linewidth}
        \includegraphics[width=0.32\linewidth]{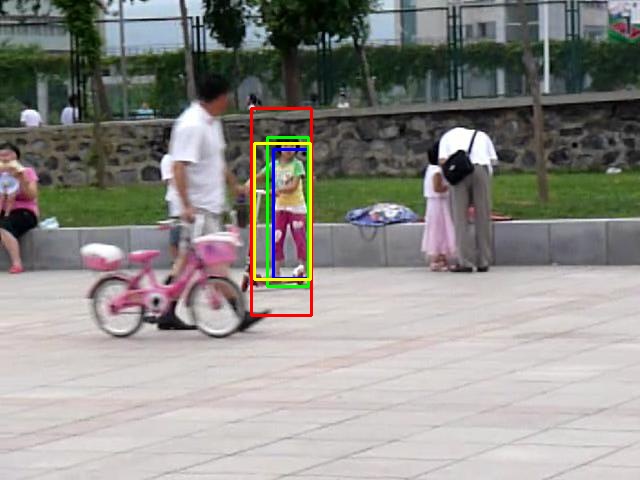}
        \includegraphics[width=0.32\linewidth]{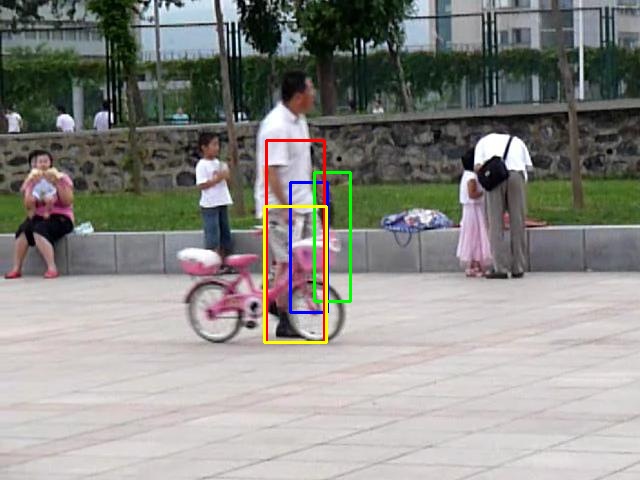}
        \includegraphics[width=0.32\linewidth]{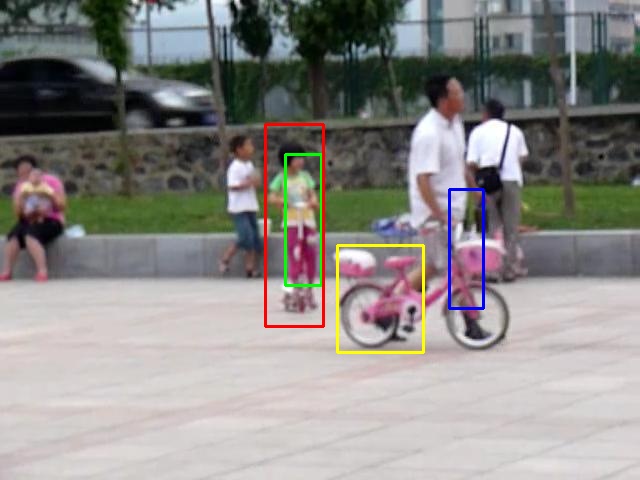}
        \caption{{\it girl}: total occlusion}
    \end{subfigure}
    \begin{subfigure}{0.96\linewidth}
        \includegraphics[width=0.32\linewidth]{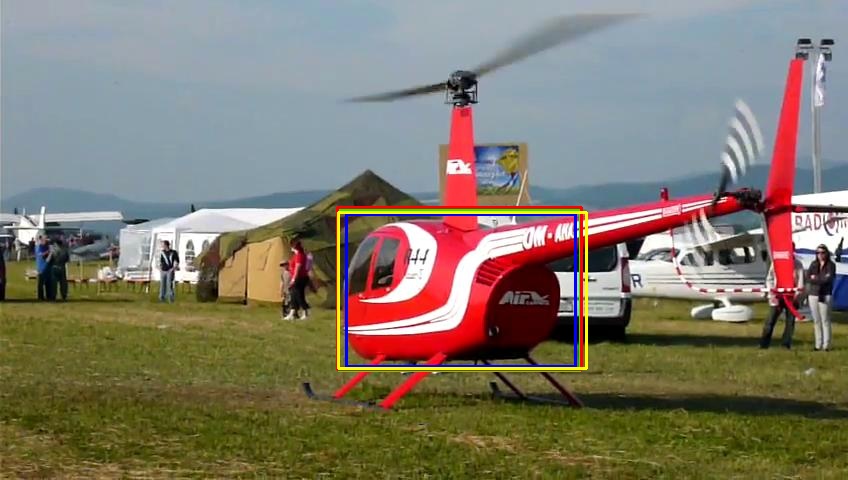}
        \includegraphics[width=0.32\linewidth]{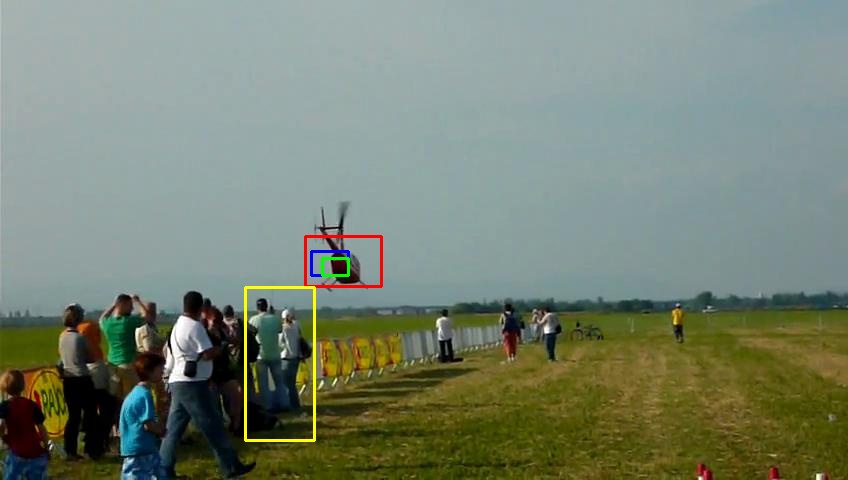}
        \includegraphics[width=0.32\linewidth]{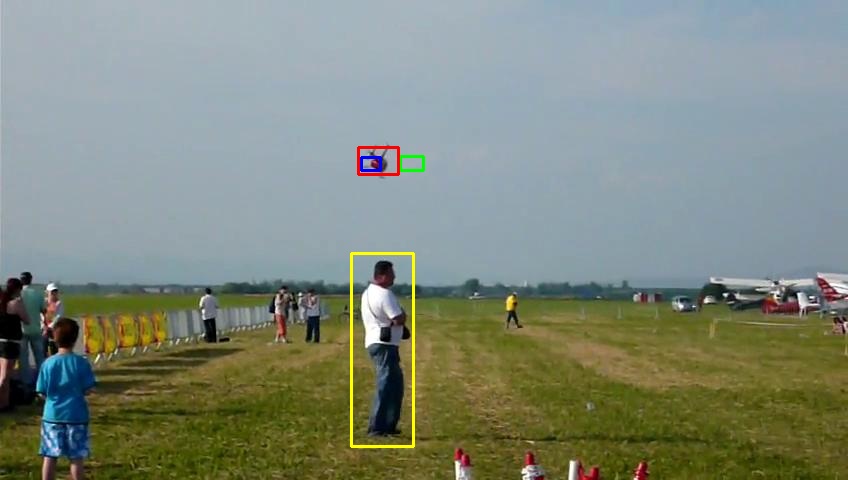}
        \caption{{\it helicopter}: abrupt changes in target appearance}
    \end{subfigure}
    \begin{subfigure}{0.96\linewidth}
        \includegraphics[width=0.32\linewidth]{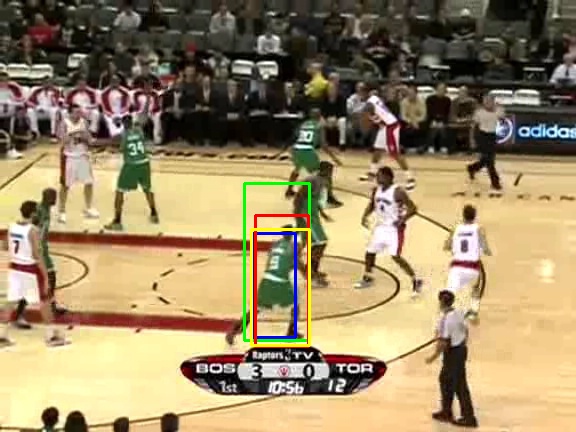}
        \includegraphics[width=0.32\linewidth]{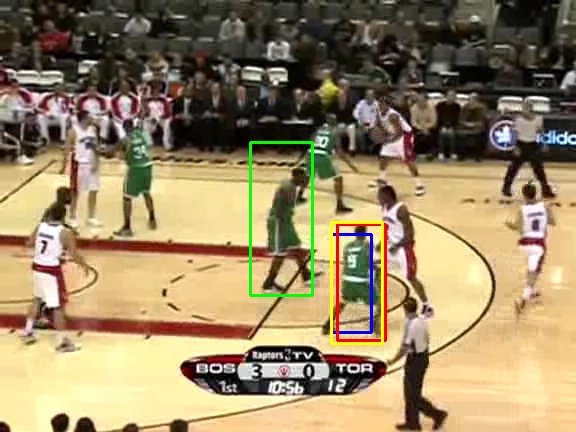}
        \includegraphics[width=0.32\linewidth]{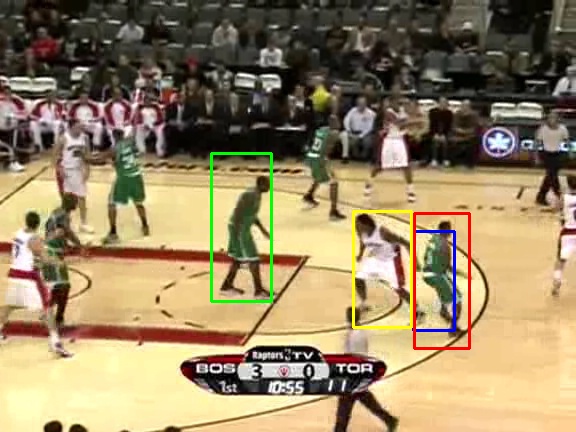}
        \caption{{\it basketball}: multiple object instances}
    \end{subfigure}
    \begin{subfigure}{0.96\linewidth}
        \includegraphics[width=0.32\linewidth]{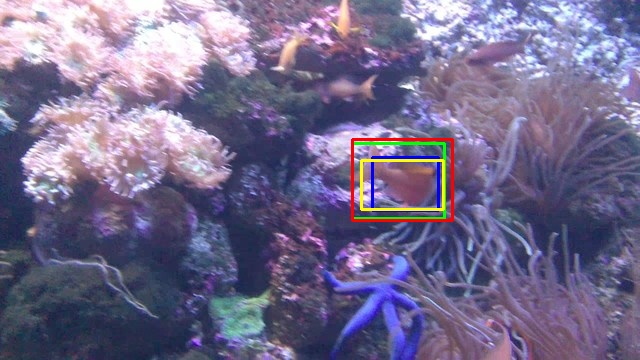}
        \includegraphics[width=0.32\linewidth]{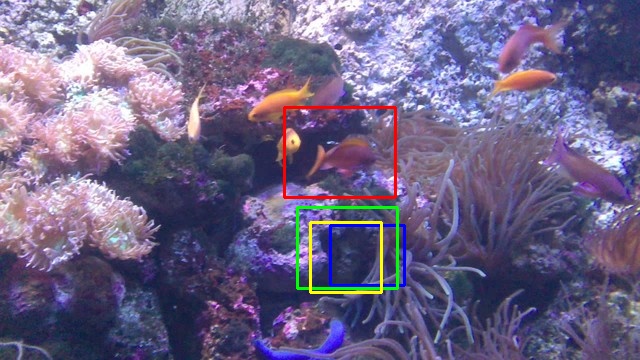}
        \includegraphics[width=0.32\linewidth]{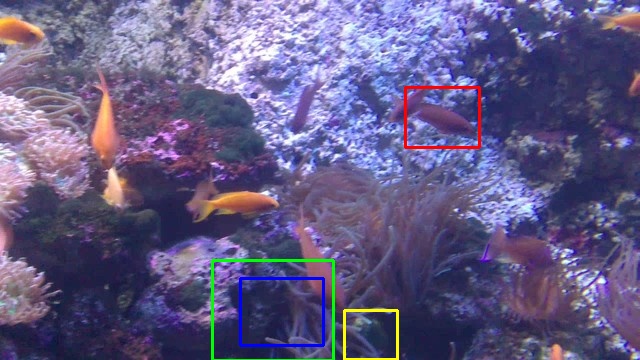}
        \caption{{\it fish}: similar foreground and background features}
    \end{subfigure}
\caption{DAWN can handle various difficult situations in unsupervised video object tracking whereas existing methods (ECO~\cite{eco}, ADNet~\cite{adnet}, DaSiamRPN~\cite{dasiamrpn}) fail in at least one of them.}
    \label{fig:teaser}
\vspace{-0.1in}
\end{figure}

Despite recent progress in deep learning and pertinent research on video object tracking, as exemplified in VOT challenges~\cite{vot2016,vot2017} and OTB benchmarks~\cite{otb2013,otb2015}, top contenders in these challenges and benchmarks still have problem in unsupervised object tracking where for a given video only the bounding box in the first frame is given and no reset is allowed, in the presence of total occlusion, abrupt changes in target appearance, multiple object instances, and similar foreground and background features, see Figure~\ref{fig:teaser}. 

In this paper, we introduce a deep neural network augmented with both foreground and background memory blocks which are updated by an attention LSTM as a memory controller for unsupervised video object tracking.  Central to our network architecture is an external memory module to remember (and forget) evolving target object and scene structures which are inspired by~\cite{mavot,memtrack}, thus setting such memory-augmented tracking methods apart from top tracking methods such as~\cite{mdnet,deeptrack} (online learning),~\cite{adnet,DRL,rdt} (deep reinforcement learning), and \cite{eco,ccot} (coupling deep features with traditional features).

The central issue in unsupervised video object tracking is that target appearance given in the first frame is likely to dramatically change during the course of tracking, which can be complicated by occlusion and image degradation such as motion blurs.  Online tracking is a typical method to adjust a given model to adapt to changes during tracking.  However, updating network parameters through online backpropagation can be quite computationally expensive, as opposed to updating an external memory block where a lightweight read and write strategy can be adopted. 

Such memory updating strategy for unsupervised object tracking, on the other hand, is susceptible to problems such as total occlusion and severe motion blurs which can contaminate memory blocks. To protect the memory from erroneous update due to the above problems, we utilize an attention LSTM as a memory controller.  The attentional module attenuates the effect of noise by focusing on the target despite partial occlusion or similar background features, thus making the memory I/O less vulnerable in our method. 

By adding an identical memory module to remember (and forget) the evolving {\em background} as well, our dual memory augmented network (DAWN) can suppress confusing background that may look similar to the tracking target, and thus can seamlessly deal with the above difficult tracking situations which are problematic to conventional trackers. Such confusing background can distract these previous models and make them start tracking the occluder and thus the wrong object.

To  focus on dual memory and attention LSTM, we fix the aspect ratio of our bounding boxes on DAWN to make us comparable to previous memory-based methods~\cite{memtrack,mavot}. To  eliminate contributing factors other than the two technical contributions, we do not have sophisticated engineering other than restarting DAWN after total occlusion, so we evaluate our model in VOT2016~\cite{vot2016} and VOT2017~\cite{vot2017} competitions including comparison with the VOT2018~\cite{vot2018} real-time champion, showing DAWN performs significantly better than state-of-the-art trackers while running fps $>$ 10. 
%\textcolor{blue}{Note that VOT2018 used the same dataset to test real-time trackers.} 

%------------------------------------------------------------------------
\section{Related work}

%-------------------------------------------------------------------------
\subsection{Deep Feature in Tracking}
Over the past decade, we have witnessed the significant development of deep learning on many important computer vision tasks, particularly object classification and detection~\cite{vgg,resnet,faster_rcnn}. Recently deep learning has been extensively employed in tracking thanks to its outstanding feature representations. A number of trackers, coupling deep features with traditional features as correlation filters such as C-COT~\cite{ccot} and ECO~\cite{eco} have shown their effectiveness on various tracking benchmarks~\cite{otb2013,otb2015} and challenges~\cite{vot2016,vot2017}. 

Tracking methods based on Tracking-by-Detection~\cite{survey,survey2,deeptrack} build a classifier that separates the target from the enclosing background. However, training data of these online-only approaches is the input video itself which fundamentally limits their learned models. The MDNet~\cite{mdnet} solves this problem by end-to-end learning of an offline deep feature extractor, where the classifier will be refined online. Although it has achieved competitive performance, its speed is restricted by online training and updating. 

Deep feature matching based methods are gaining much attention because of its excellent speed and tracking performance. The GOTURN~\cite{goturn} learns target tracking states by comparing and matching feature pairs in consecutive frames, so it cannot handle target occlusion in principle. The SiamFC~\cite{SiamFC} uses fully convolutional Siamese networks to match features between template frame and target frame. However, this method only utilizes the first frame as the template which will not work well when the tracking object undergoes notable deformation in the long term. 

\begin{figure*}[tb]
    \centering
    \includegraphics[width=0.96\linewidth]{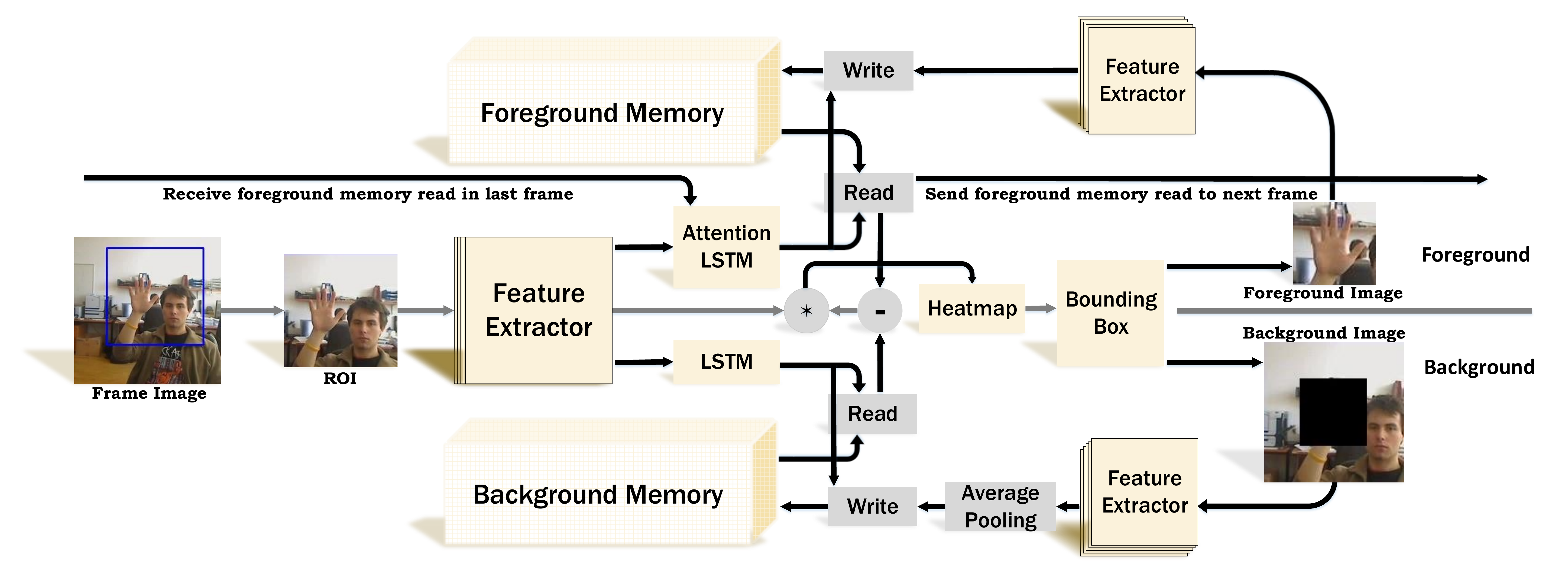}
    \caption{General structure of our network, where {\bf attention LSTM} (Figure~\ref{fig:attention_comparison}) and {\bf dual memory} (Figure~\ref{fig:dual_mem}) will be detailed. Blue rectangle in the frame image is the Region Of Interest (ROI) where ROI features are extracted. ROI features are then fed into attention LSTM together with foreground memory read in the last frame, which are also fed into a conventional LSTM. The outputs from two LSTMs are used to read memories from foreground and background memory blocks respectively. These memories are then used to generate a heat map to predict the bounding box which divides ROI into two parts, a foreground image, and a background image. Features extracted from the two images are then used to write into foreground and background memory blocks respectively.}
    \label{fig:model}
\vspace{-0.2in}
\end{figure*}

%-------------------------------------------------------------------------
\subsection{Memory Structure in Tracking}
To solve the static template frame problem in GOTURN and SiamFC mentioned above, memory blocks are adopted in a number of contemporary trackers. The RDT~\cite{rdt} uses a template pool to store the target's different appearances. Even though its template pool updating is trained by deep reinforcement learning~\cite{DRL}, the target appearances written and read are discrete. MemTrack~\cite{memtrack} uses NTM (Neural Turing Machine)~\cite{NTM} as its memory block to store continuous feature of the target, and uses LSTM~\cite{lstm} to dynamically control memory reading and writing. They used gated residual template learning to manage the quantity of retrieved memory which is used to mix with the starting template. This memory architecture is commonly used in many fields of deep learning, especially in one-shot learning, such as~\cite{mmnet,santoro2016one}. On the other hand, MAVOT~\cite{mavot} retrieves and memorizes the information of both foreground and background to support background suppression. A similar idea was adopted in DSiam~\cite{DSiam} where a dynamic Siamese network was used for target image variation and background suppression. Our improved attention LSTM memory controller is closely related to~\cite{memtrack} and background memory to~\cite{mavot}, which will be explained in the next section.

%-------------------------------------------------------------------------
\subsection{Attention Mechanism in Tracking}
Attention models, initially applied in image recognition tasks, was then combined with a recurrent model to form Recurrent Attention Model (RAM)~\cite{RAM}. The RAM soon became popular in multiple areas such as image captioning~\cite{Xu2015ShowAA}, image classification~\cite{Wang_2017}, pose estimation~\cite{Chu_2017_pose}, etc. In multiple objects tracking, STAM~\cite{Chu_2017_attention} uses the attention model to deal with occlusions and interactions among targets. For single object tracking, RASNet~\cite{RASNet} introduces general attention, residual attention and channel attention. MemTrack~\cite{memtrack} uses a common RAM structure similar to those in NLP tasks to roughly locate the object, and provide the memory block a correct retrieval key. However, we found that this structure can give bad location estimation and consequently, it tracks the background and memorizes the relative location of the object context. In this paper, we modify the RAM structure to produce more accurate estimations of the target's location. While a single object is tracked in unsupervised, it may be partially or totally occluded during tracking, which will cause the model especially one with memory blocks to track another object instead of the original target. Our attention mechanism is designed to solve the occlusion problem, which demonstrates more robust behavior than memory structures without attention mechanism such as~\cite{mavot}.

%------------------------------------------------------------------------
\section{Dual Augmented Memory Network}
We present our DAWN model in detail. Figure~\ref{fig:model} shows the overall framework: our model consists of a dual memory structure for remembering the evolving foreground and background features, an improved attention LSTM and a conventional LSTM for controlling the read/write operations of the foreground and background memory respectively. Note that {\em no} backpropagation is run, online or offline during the tracking process. All feature extractors in DAWN share the same weights. In this section, $*$, $\cdot$ and $\odot$ denote element-wise multiplication, matrix multiplication and convolution operation respectively.

%-------------------------------------------------------------------------
\subsection{Feature Extraction}
We follow the Fully Convolutional Neural Networks structure from SiamFC~\cite{SiamFC} to extract deep features within the ROI which include the deep features for both target and background. The output will then be fed into three branches, namely, ROI branch, foreground branch, and background branch. The ROI branch receives an ROI patch from the newest frame and bounding box predicted in the last frame. The output of the ROI branch is an $n \times n \times c$ matrix $\mathbf{F}$,  which consists of vectors $\mathbf{f}_{i,j}$ of size $c$ where $0 \leq i,j < n$. Working with the memory I/O, $\mathbf{F}$ will be used to predict the new bounding box. 

The foreground branch receives the newest target patch from the bounding box predicted in ROI branch. The output of the foreground branch is an $m \times m \times c$ matrix $\mathbf{F_{fore}}$, which will be written to foreground memory. 

The background branch receives the background information in ROI and outputs the feature of the background branch, an $n \times n \times c$ matrix, which will be written in background memory after an additional average pooling layer. 
%-------------------------------------------------------------------------
\vspace{-0.18in}
\subsection{Attention Based on Foreground Memory}
\label{sec:attention}
\vspace{-0.07in}
In video object tracking, a bounding box is an axis-aligned rectangle, so it includes a lot of background regions as well. Thus features extracted from a given bounding box contain noises which adversely affect both the performance and robustness of the model. To extract object features with less noise, Yang and Chan (MemTrack)~\cite{memtrack} introduced an attention LSTM as a memory controller, but their attention module acts more like a simple protocol for memory reading and writing. Often times, their attention seems to be random and consequently the model tracks some part of the background and remembers the target's relative position with respect to the tracked background regions.  Thus, once the target's relative position starts to change, MemTrack will lose track, see Figures~\ref{fig:handnbook} and~\ref{fig:butt}. 

\begin{figure}[htb]
    \centering
%\begin{tabular}{cc}
    \begin{subfigure}{0.70\linewidth}
        \includegraphics[width=\linewidth]{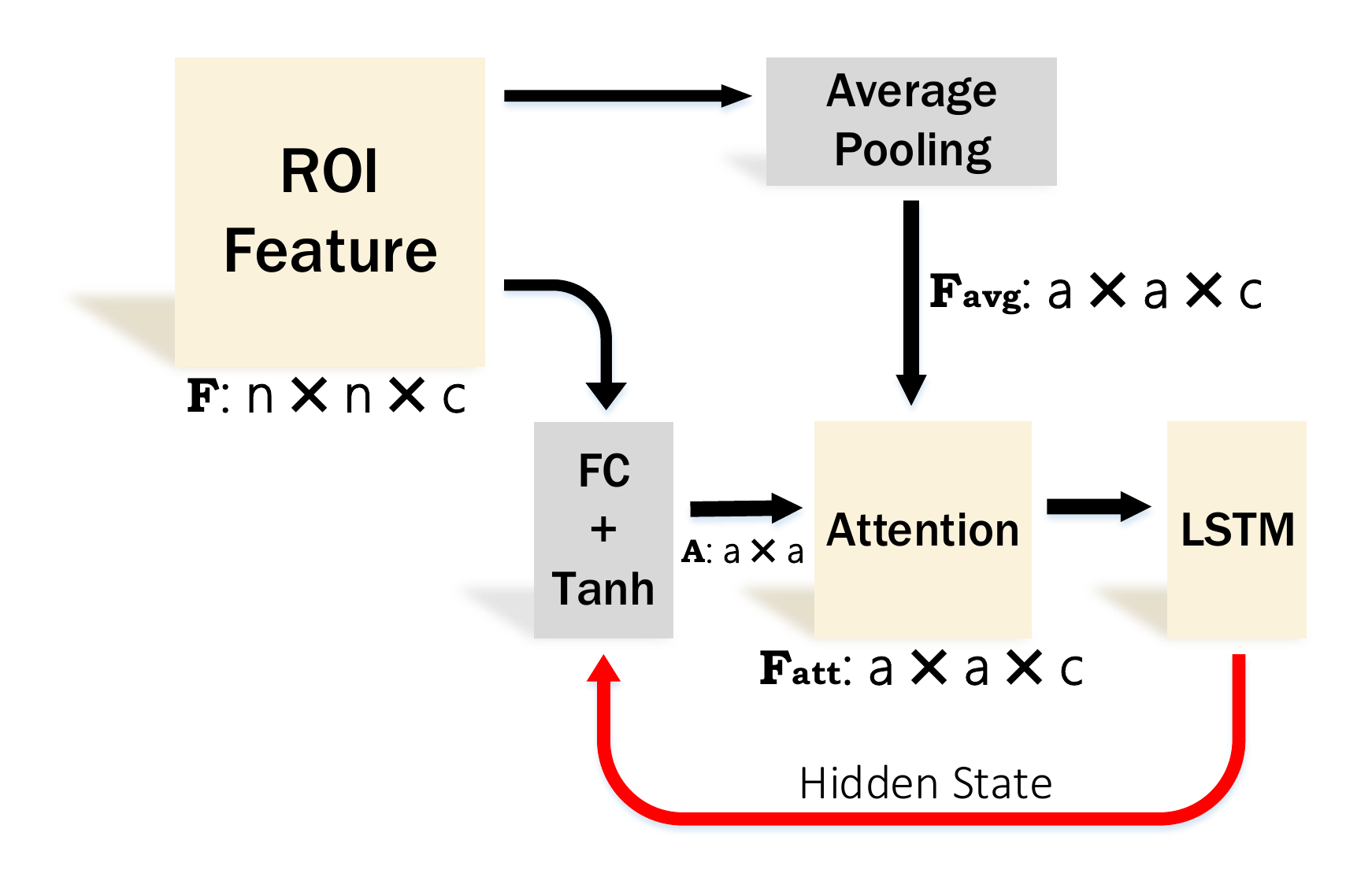}
        \caption{MemTrack}
    \end{subfigure} \\
    \begin{subfigure}{0.70\linewidth}
        \includegraphics[width=\linewidth]{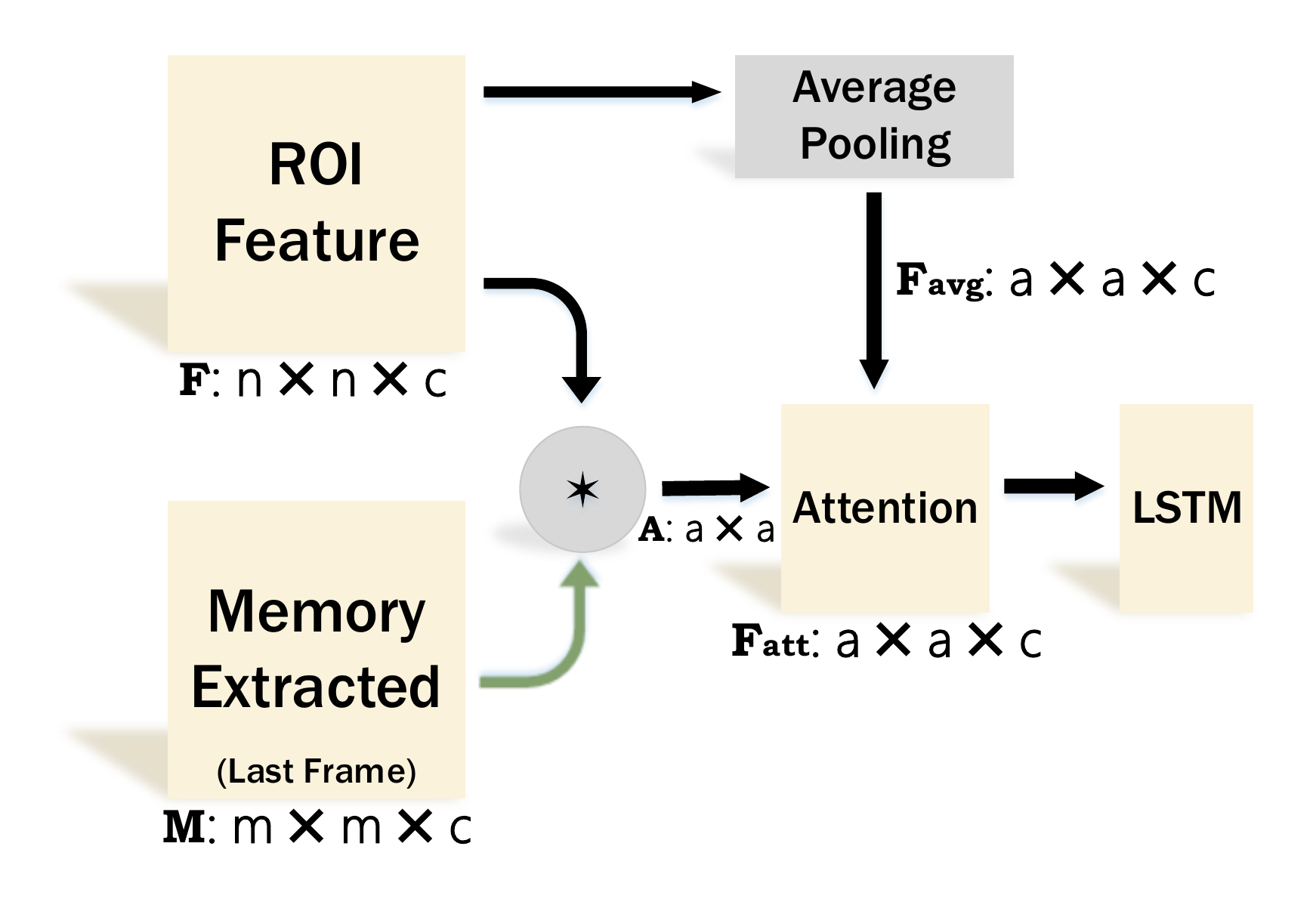}
        \caption{DAWN}
    \end{subfigure}
%\end{tabular}
    \caption{Attention modules of MemTrack~\cite{memtrack} and DAWN.}
    \label{fig:attention_comparison}
\vspace{-0.05in}
\end{figure}

Since attention can be roughly interpreted as an estimation of the target object's position, the concept of memory augmentation can also be applied here. Specifically, to achieve better estimation, we introduce a memory augmented attention block whose difference from the attention block in MemTrack~\cite{memtrack} is shown in Figure~\ref{fig:attention_comparison}. In DAWN, we utilize the memory read from the last frame, i.e., a $m \times m \times c$ matrix $\mathbf{M}$ consists of vectors $\mathbf{m}_{i,j}$ of size $c$, where $0 \leq i,j < m$.

First, to generate attention score, $\mathbf{M}$ is convolved with $\mathbf{F}$ which outputs an $a \times a$ matrix $\mathbf{A}$, as weights of attention scalars $a_{i,j}$, where $0 \leq i,j < a$ and $a = n - m + 1$, whose value is given by
\begin{equation}
    a_{i,j} = \frac{\exp(r_{i,j})}{\displaystyle\sum_{s=0}^{a-1}\displaystyle\sum_{t=0}^{a-1} \exp(r_{s,t})},
\end{equation}
where
\begin{equation}
    r_{i,j} = \displaystyle\sum_{s = 0}^{m-1} \displaystyle\sum_{t = 0}^{m-1} \mathbf{f}^{\top}_{i + s,j + t} \cdot \mathbf{m}_{s,t}.
\end{equation}

Then, we use average pooling on $\mathbf{F}$ with filter size $m \times m$ and stride size 1, and obtain $a \times a \times c$ matrix $\mathbf{F_{avg}}$:
\begin{equation}
    \mathbf{F_{avg}} = \mathit{AveragePooling}_{m \times m}(\mathbf{F}).
\end{equation}

Finally, we use the attention score $\mathbf{A}$ and averaged feature $\mathbf{F_{avg}}$ to produce attention feature $\mathbf{F_{att}}$:
\begin{equation}
\mathbf{F_{att}} = \mathbf{A} * \mathbf{F_{avg}}.
\end{equation}

Comparing with MemTrack~\cite{memtrack}, our attention mechanism has the following advantages: 
\begin{enumerate}
\setlength{\itemsep}{0.0ex}
\item We use the last frame to extract the memory for generating attention, more inter-frame information being incorporated in DAWN during tracking. This is a more direct approach than MemTrack~\cite{memtrack}, which uses the LSTM hidden layer and current ROI feature to generate attention.
\item Our mechanism allows us to use the first frame target feature to initialize the memory, which is more stable and robust than LSTM hidden layer initialization. See the first frame in Figure~\ref{fig:butt}.
\item The uncompressed memory extracted from the last frame has less information loss than the compressed LSTM hidden layer.
\end{enumerate}

%-------------------------------------------------------------------------
\subsection{Memory Controller}
\label{sec:controller}
DAWN uses standard LSTM with layer normalization~\cite{layer} and dropout regularization~\cite{dropout} in both foreground and background memory controllers, see Figure~\ref{fig:dual_mem}. The input to foreground and background memory controller are $\mathbf{F_{att}}$ and $\mathbf{F_{avg}}$ respectively, computed in section~\ref{sec:attention}. With the input feature and the previous hidden state $\mathbf{h_{t-1}}$, the hidden state updates to $\mathbf{h_{t}}$ in both controllers. Our memory reading and writing process are inspired by MemN2N~\cite{memn2n} and MemTrack~\cite{memtrack}. Here, we describe foreground memory reading and writing process. The background memory read/write is similar. 

During a reading process, a template is retrieved from all the memory slots as a weighted sum, where the read weights are the \textit{softmax} result of the cosine similarity between the given read key $\mathbf{r_{t}}$ and all the memory keys. The read key $\mathbf{r_{t}}$ is calculated by hidden states $\mathbf{h_{t}}$ of LSTM:
\begin{equation}
\mathbf{r_{t}} = \mathbf{W}^{r} \cdot \mathbf{h_{t}} + \mathbf{b}^{r},
\end{equation}
where $\mathbf{W}^{r}$ and $\mathbf{b}^{r}$ are weight matrix and bias.
The memory key is the output of $m \times m$ average pooling on the corresponding memory slot. 

During a writing process, the write weight $\mathbf{w_{t}}$ controls whether to update the memory and which slots to update. Using hidden states $\mathbf{h_{t}}$, we apply the same method as MemTrack~\cite{memtrack} to generate $\mathbf{w_{t}}$. The writing process is defined by 
\begin{equation}
\mathbf{M_{t}} = (1-\mathbf{w}_{t}) * \mathbf{M_{t-1}} + \mathbf{w}_{t} * \mathbf{F_{fore}},
\end{equation}
where $\mathbf{F_{fore}}$ is the output of the foreground branch.

For LSTM initialization, we use the initial target's feature with average pooling and~\textit{tanh} activation to generate hidden state $\mathbf{h_{0}}$ and cell state $\mathbf{c_{0}}$.  We apply the Residual Template Learning adopted in MemTrack~\cite{memtrack} in foreground memory but not in background memory.

%-------------------------------------------------------------------------
\subsection{Dual Memory Structures} \label{sec:DMS}
\begin{figure}[htb]
    \centering
    \includegraphics[width=0.80\linewidth]{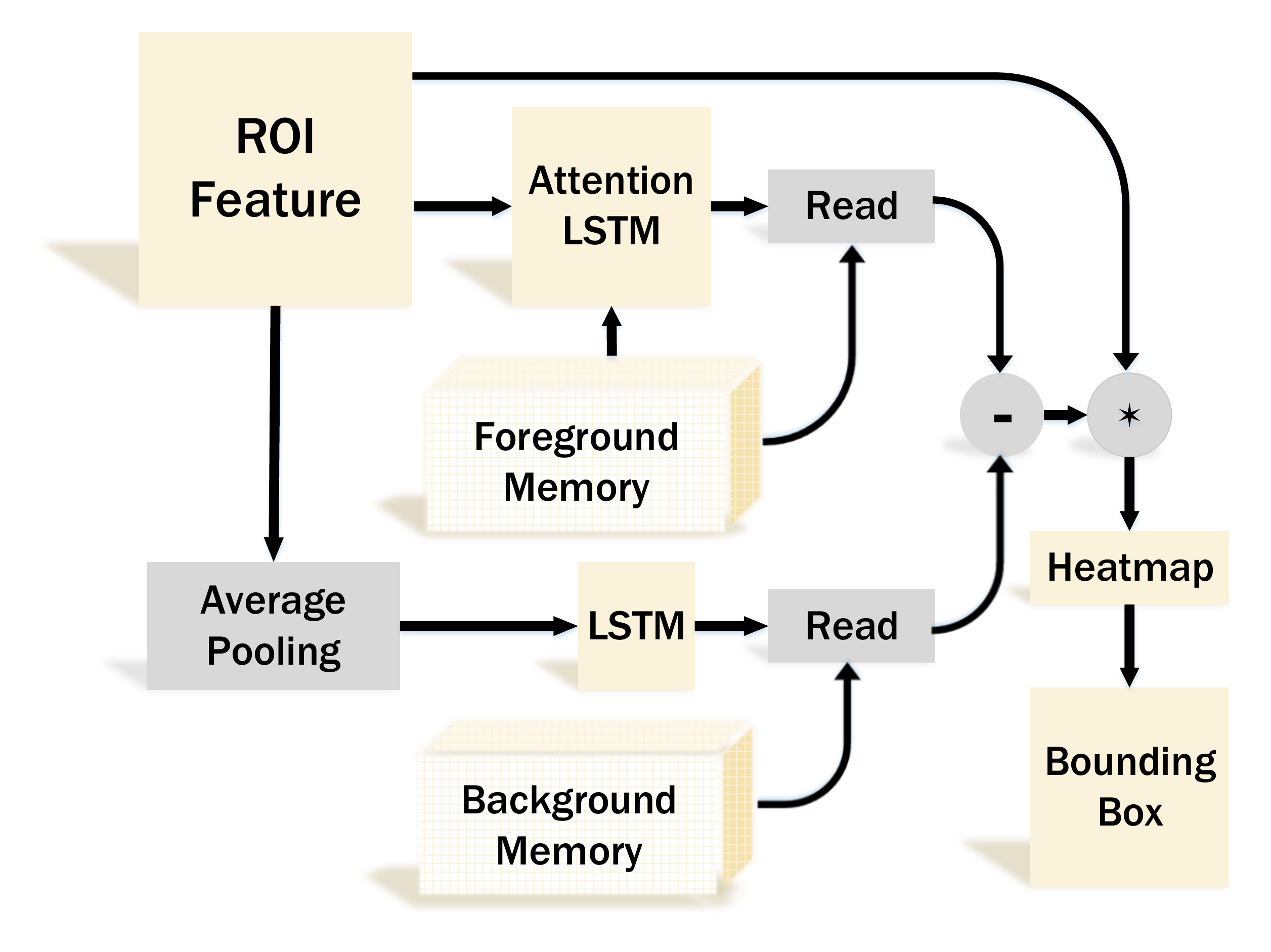}
    \caption{Dual Memory Structure of DAWN.}
    \label{fig:dual_mem}
\vspace{-0.15in}
\end{figure}

Since memory structure with foreground-only features shows limitations in tackling problems such as occlusion, multiple objects, and similar background features, etc., DAWN uses dual memory blocks, see Figure~\ref{fig:dual_mem}. Foreground memory can remember the appearance of the target, while background memory can record surrounding variation.

Reading from foreground and background memory has been discussed in Section~\ref{sec:attention} and~\ref{sec:controller}. After reading foreground memory $\mathbf{M}$ and background memory $\mathbf{M_{back}}$,  both with dimension $m \times m \times c$, we obtain $\mathbf{\overline{M}}$ by non-maximal suppression using subtraction:
\begin{equation}
\mathbf{\overline{M}} = \mathbf{M} - \mathbf{M_{back}}
\end{equation}

$\mathbf{\overline{M}}$ is convolved with the $\mathbf{F}$ to obtain the heat map $\mathbf{H}$ with dimension $a \times a$, which is calculated by:
\begin{equation}
    \mathbf{H} = \mathbf{F} \odot \mathbf{\overline{M}}.
\end{equation}

We up-sample the heat map $\mathbf{H}$ as in SiamFC~\cite{SiamFC} to predict the bounding box. Then we extract features from the image cropped with the bounding box and write them to the corresponding memory blocks. An additional average pooling layer is applied before writing background features into background memory.

Comparing with memory blocks only involving foreground features~\cite{memtrack}: 
\begin{enumerate}
\setlength{\itemsep}{0.0ex}
\item {\em Occlusion}. DAWN can suppress the adverse influence of possible memory contamination caused by occlusion. Previously, when the target is being occluded, features of the occluding object will be written into and thus contaminate foreground memory (see `girl' and `frisbee' in Figure~\ref{fig:mem}), consequently causing the model to track a false background object. But with DAWN's background memory block, since the memory of a background object has been recorded in previous frames, subtracting its memory can mitigate the above contamination while keeping track the target after occlusion. When detecting total occlusion, DAWN will not update bounding box until the target re-appears in a subsequent ROI. 
\item {\em Similar Background}. DAWN can distinguish other similar background objects or features, which usually confuse trackers with only foreground memory.  Since background features have been memorized, although they look similar to the target in foreground memory, DAWN can still differentiate them after subtracting the relevant background memory. Using cosine window to penalize large displacements can suppress similar objects in the background. See `basketball', `fish' in Figure~\ref{fig:teaser}, and `godfather', `gymnastics' in Figure~\ref{fig:mem}.
\end{enumerate}

\subsection{DAWN-PRN: DAWN with Region Proposal Network
}
\begin{figure}[htb]
    \centering
    \includegraphics[width=0.96\linewidth]{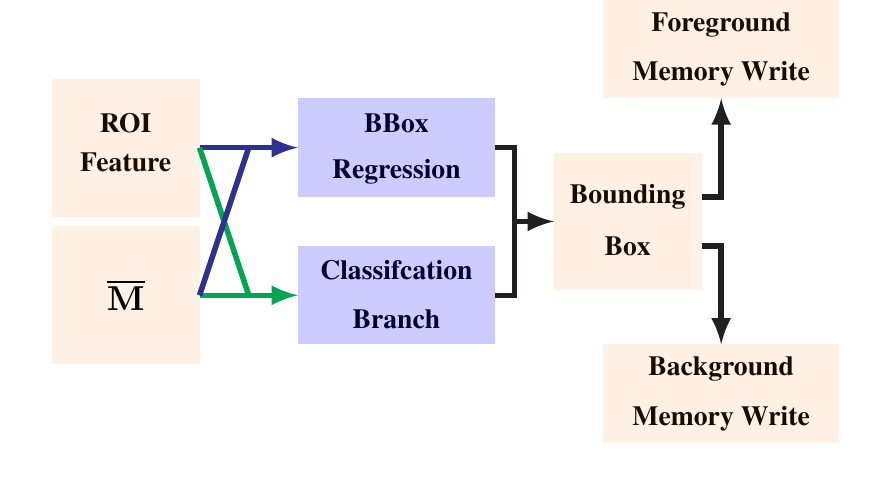}
    \caption{
    DAWN-PRN: $\overline{M}$ is same and defined by Eq.~(7).}
    \label{fig:model_rpn}
% \vspace{-0.15in}
\end{figure}

We change our tracking backbone from SiamFC~\cite{SiamFC} to SiamRPN~\cite{SiamRPN}, and call it DAWN-RPN, which is shown in Figure~\ref{fig:model_rpn}. In DAWN-RPN, we adopt the techniques in SiamRPN~\cite{SiamRPN} to generate bounding boxes. The reading and writing operations of memory is unchanged.

%------------------------------------------------------------------------
\section{Experiments}
We will present the comparative results with other trackers using VOT toolkits on two challenging tracking competitions: VOT2016 and VOT2017, and then show the respective improvement due to the attention and memory module. Please see our videos at \emph{https://zhmeishi.github.io/DAWN/} .

\begin{figure*}[t]
    \centering
    \includegraphics[width=0.10\linewidth]{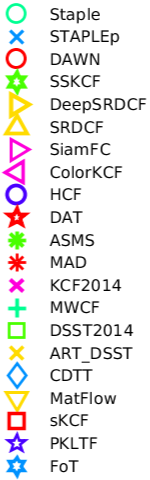}
    \includegraphics[width=0.34\linewidth]{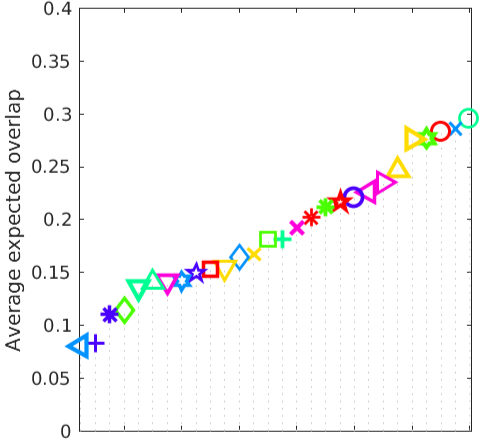}
    \includegraphics[width=0.10\linewidth]{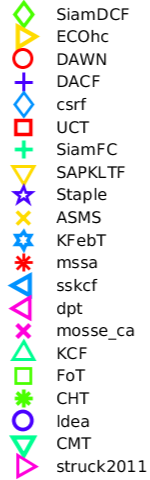}
    \includegraphics[width=0.34\linewidth]{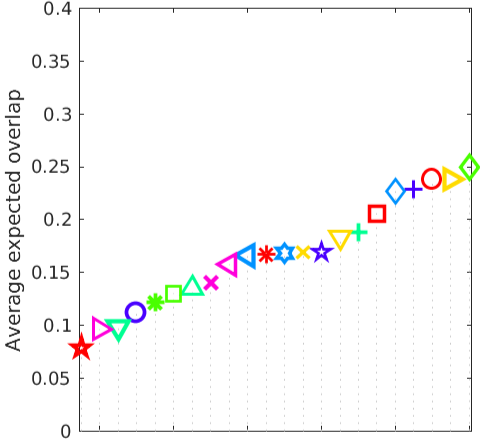}
    \caption{DAWN is ranked $3^{rd}$ in VOT2016 (left) and VOT2017 (right) among all the VOT fast trackers running at fps $>$ 10. }
    \label{fig:stat1617}
\vspace{-0.15in}
\end{figure*}

\subsection{Implementation Details}
We apply the same Alex-like~\cite{alexnet} CNN feature representation as SiamFC~\cite{SiamFC}. The input image sizes of the ROI, target and background branch are respectively $255 \times 255 \times 3$, $127 \times 127 \times 3$ and $255 \times 255 \times 3$. The output matrix size of the foreground branch is $6 \times 6 \times 256$, and, for the ROI branch and background branch, the size is $22 \times 22 \times 256$. Our DAWN is pre-trained offline on the video object detection dataset of ImageNet Large Scale Visual Recognition Challenge (ILSVRC15)~\cite{imagenet}. The offline training strategy is the same as MemTrack~\cite{memtrack}, where the training optimizer is Adam~\cite{adam} with initial learning rate $1e^{-4}$ and the weight decay set to be $0.02$ every $10,000$ iterations. The number of memory slots is 8. We suppress the heat map with a cosine window by an exponential factor of $0.27$. The target image size is $1.32$ times bounding box size. We update the target size with exponential smoothing  $s_{t} = (1 - \alpha )s_{t-1} + \alpha s_{new}$ to deal with scale changes, where $s$ is the target size and the exponential factor $\alpha$ is 0.6.

Our tracker is implemented in Python using the TensorFlow framework. All of the experiments were conducted on the following hardware specifications: Intel(R) Core(TM) i7-7700HQ CPU @ 2.80GHz, 16 GB RAM, and NVIDIA GPU GTX1070.

\subsection{VOT2016 and VOT2017 Results}

Using the VOT toolkit, we compare DAWN with all fast trackers (fps $> 10$) in VOT2016 and VOT2017 competitions which respectively contains 60 video sequences. Figure~\ref{fig:stat1617} shows the comparison results. Only fine-tuned on the training dataset, DAWN has consistently good performance in VOT2016 and VOT2017 with same hyper-parameters. Though DAWN ranks the third in VOT2016 and VOT2017, where the rank is measured by the expected overlap ratio (EAO), but our bounding box aspect is fixed, DAWN still has an excellent success rate in both VOT2016 (Table~\ref{table:suc_rate_16}) and VOT2017 (Table~\ref{table:suc_rate_17}) with the least number of failures.
\vspace{-0.1in}
\begin{table}[!h]
  \centering
\begin{tabular}{l|ccc}
\hline
      Measures & DAWN  & STAPLEp & Staple\\
\hline
EAO ($\uparrow$)           & \textcolor[rgb]{0,1,0}{0.28} & \textcolor[rgb]{0,0,1}{0.29} & \textbf{\textcolor[rgb]{1,0,0}{0.30}} \\
Failures ($\downarrow$)      &
\textbf{\textcolor[rgb]{1,0,0}{18.58}} & \textcolor[rgb]{0,1,0}{24.32} & \textcolor[rgb]{0,0,1}{23.89} \\
\hline
\end{tabular}
\caption{Model Performance in VOT2016}
\label{table:suc_rate_16}
\vspace{-0.15in}
\end{table}
\vspace{-0.1in}
\begin{table}[!h]
  \centering
\begin{tabular}{l|ccc}
\hline
Measures & DAWN  & ECOhc & SiamDCF\\
\hline
EAO ($\uparrow$)           & \textcolor[rgb]{0,0,1}{0.24} & \textcolor[rgb]{0,0,1}{0.24} & \textbf{\textcolor[rgb]{1,0,0}{0.25}} \\
Failures ($\downarrow$)      &
\textbf{\textcolor[rgb]{1,0,0}{26.58}} & \textcolor[rgb]{0,0,1}{28.77} & \textcolor[rgb]{0,1,0}{29.41} \\
\hline
\end{tabular}
\caption{Model Performance in VOT2017}
\label{table:suc_rate_17}
\vspace{-0.1in}
\end{table}

\subsection{VOT unsupervised Tracking}\label{sec:unsupervised}

Here we choose videos that are very difficult to track in the unsupervised mode and provide brief information as well as major tracking difficulties. Then we will tabulate the tracking results of DAWN and current state-of-the-art trackers.

The following summarizes the major problems in unsupervised tracking with their abbreviations used in Table~\ref{result_table}:
\vspace{-0.2in}
\begin{table}[!h]
  \centering
\begin{tabular}{ll}
OC: & Occlusion (Total or Partial) \\
MB: & Motion Blur (Severe) \\
AC: & Abrupt Changes in Target Appearance \\
SB: & Similar Background Objects or Features \\
\end{tabular}
\vspace{-0.1in}
\end{table}

\begin{table*}[th]
  \centering
  \linespread{0.0}
\begin{tabular}{l|c|ccc|ccc|ccc}
\hline
Video       & DAWN & MemTrack & MAVOT & SiamFC & KCF & DSST & DaSiam & ADNet & C-COT & ECO \\
\hline
\footnotesize{wiper \hfill(OC, MB)}         & \checkmark & \checkmark &  & \checkmark & \checkmark & \checkmark &  & \checkmark & \checkmark & \checkmark \\
\footnotesize{bolt1 \hfill(SB)}          & \checkmark &  &  & \checkmark & \checkmark &  & \checkmark & \checkmark & \checkmark & \checkmark \\
\footnotesize{butterfly \hfill(AC, SB)}         & \checkmark &  & \checkmark & \checkmark &  &  & \checkmark & \checkmark & \checkmark & \checkmark \\
\hline
\footnotesize{helicopter \hfill(AC)}      & \checkmark & \checkmark &  & \checkmark & \checkmark & \checkmark &  & \checkmark &  & \checkmark \\
\footnotesize{godfather \hfill(OC, SB)}       & \checkmark &  & \checkmark &  & \checkmark &  &  & \checkmark & \checkmark & \checkmark \\
\footnotesize{soccer2 \hfill(OC, MB)}         & \checkmark & \checkmark &  & \checkmark &  &  & \checkmark & \checkmark & \checkmark &  \\
\hline
\footnotesize{ants1 \hfill(SB)}           & \checkmark & \checkmark &   & \checkmark &  &  &  &  & \checkmark & \checkmark \\
\footnotesize{basketball \hfill(OC, SB)}      & \checkmark &  &  & \checkmark & \checkmark &  &  &  & \checkmark & \checkmark \\
\footnotesize{motocross2 \hfill( MB,AC)}         & \checkmark &  & \checkmark &  &  &  & \checkmark &  & \checkmark & \checkmark \\
\hline
\footnotesize{ball2 \hfill(MB,SB)}         & \checkmark & \checkmark &  &  &  &  & \checkmark & \checkmark &  & \\
\footnotesize{gymnastics1 (MB, SB)}     & \checkmark &  &  &  &  &  & \checkmark &  & \checkmark & \checkmark \\
\footnotesize{girl \hfill(OC)}            & \checkmark &  & \checkmark &  &  &  &  & \checkmark &  &  \\
\hline
\footnotesize{zebrafish1 \hfill(AC, SB)}      & \checkmark &  &  &  &  &  & \checkmark &  &  & \\
\footnotesize{fish2 \hfill(AC, SB)}         & \checkmark &  &  &  &  &  &  &  &  & \\
\footnotesize{frisbee \hfill(OC, AC)}         & \checkmark &  &  &  &  &  &  &  &  & \\
\hline
fps                            & 15 & 27 & 7 & 36 & 64 & 41 & 60 & 4.5 & 0.27 & 4.6 \\
\hline
\end{tabular}
\caption{Results on state-of-the-art trackers, especially C-COT~\cite{ccot} the most precise tracker of VOT2016 and VOT2017, and DaSiam~\cite{dasiamrpn} the champion of VOT2018 real-time competition, in tracking unsupervised sequences with different technical difficulties. KCF~\cite{kcf}, DSST~\cite{dsst}, and DaSiam~\cite{dasiamrpn} are real-time trackers. \checkmark means the tracker can track the whole sequence from start to end successfully. Running speed in VOT toolkit are measured under GTX 1070. }
\label{result_table}
\vspace{-0.15in}
\end{table*}

\begin{figure}[htb]
    \centering
    \includegraphics[width=0.48\linewidth]{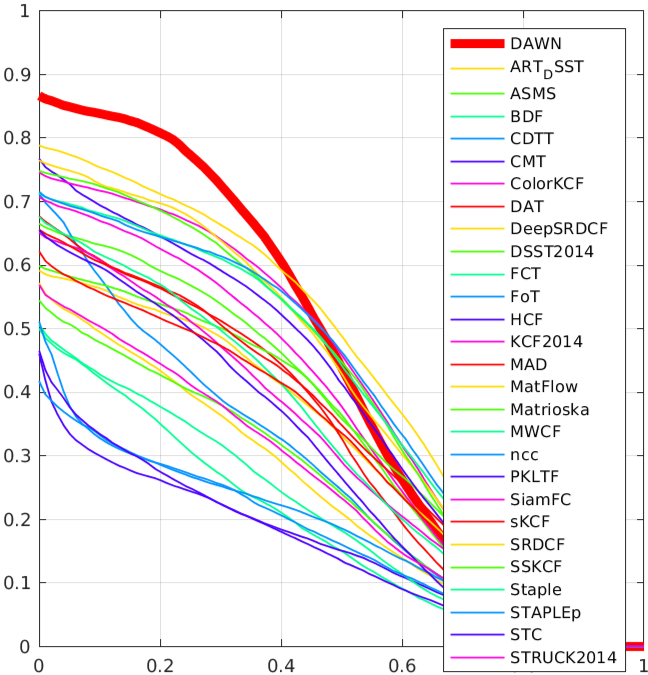}
    \includegraphics[width=0.48\linewidth]{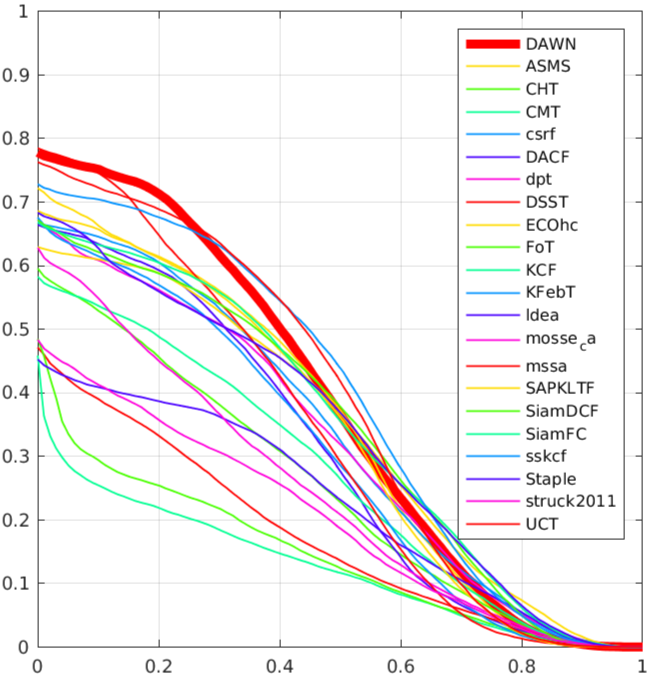}
    \caption{Comparison results on unsupervised tracking on VOT2016 (left) and VOT2017 (right). Fast trackers with fps $> 10$ in VOT toolkit are tested. In the above plots, $x$-axis indicates overlap threshold and $y$-axis accuracy.}
    \label{fig:curve1617u}
\vspace{-0.2in}
\end{figure}

Apart from a good success rate, DAWN also has excellent performance in unsupervised tracking regarding accuracy per frame as the standard. As shown in the Figure~\ref{fig:curve1617u}, DAWN enjoys an overwhelming success compared with fast trackers running at fps $>$ 10 in VOT2016, and a performance better than most fast trackers running at fps $>$ 10 in VOT2017. DAWN can focus on not losing target and the result shows that it can keep track of many hard sequences that current SOTA cannot.

\begin{figure}[!h]
    \centering
    \includegraphics[width=0.48\linewidth]{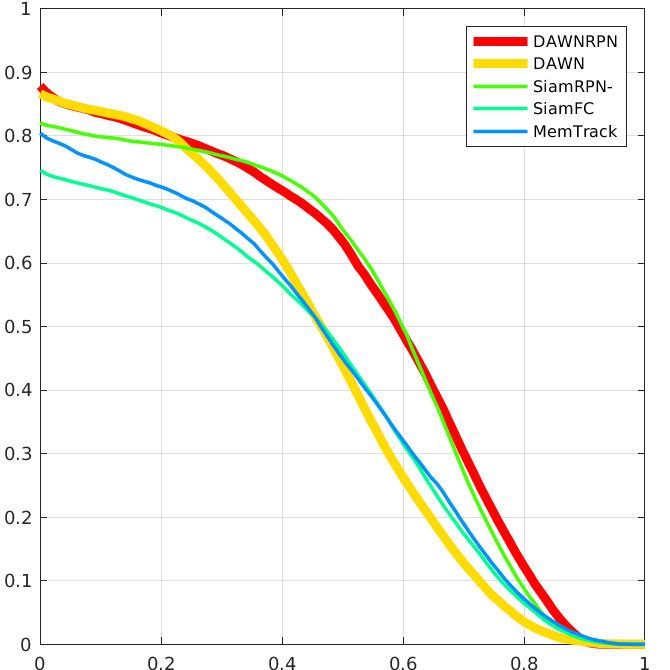}
    \includegraphics[width=0.48\linewidth]{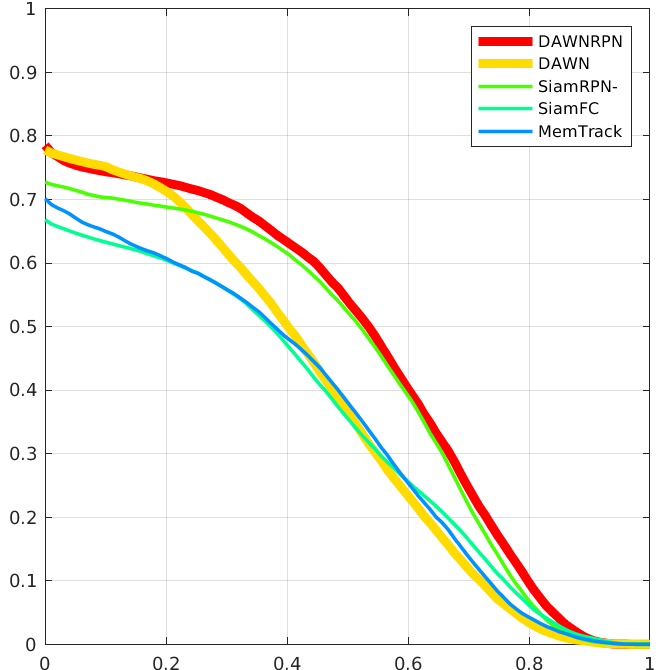}
    \caption{Unsupervised result of VOT2016 (left) and VOT2017 (right), where $x$-axis is overlap threshold and $y$-axis is accuracy.}
    \label{fig:curve}
\vspace{-0.2in}
\end{figure}

Further, we used DAWN as the pretrained model and trained the RPN branch, which was similarly done in training  SiamRPN~\cite{SiamRPN}. Then we tested DAWN-RPN on VOT2016 and VOT2017 using the VOT toolkit, and compare DAWN-RPN with DAWN and SiamRPN-- (which is our own implementation since the training code is not released). Figure~\ref{fig:curve} shows the consistent improvement  of DAWN-RPN over SiamRPN-- (and the original DAWN as well) in the unsupervised competitions in VOT2016 and VOT2017.

\subsection{Attention}

This section reports the comparison results between the attention module of DAWN and MemTrack evaluated under the VOT2016 dataset. Both models are trained on the VID dataset of ILSVRC~\cite{imagenet}.

\begin{figure}[htb]
    \centering
    \begin{subfigure}{0.98\linewidth}
        \begin{subfigure}{0.32\linewidth}
            \includegraphics[width=\linewidth]{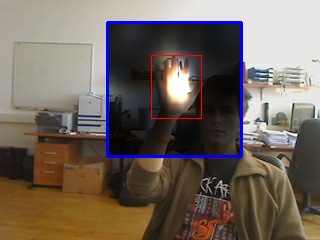}
            \includegraphics[width=\linewidth]{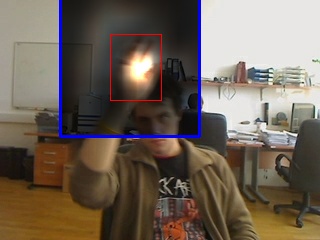}
            %\caption{DAWN}
        \end{subfigure}
        \begin{subfigure}{0.32\linewidth}
            \includegraphics[width=\linewidth]{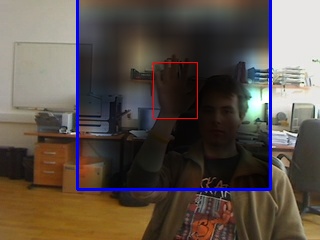}
            \includegraphics[width=\linewidth]{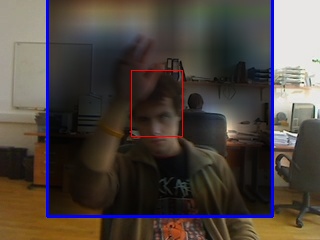}
            %\caption{Memtrack}
        \end{subfigure}
        \begin{subfigure}{0.32\linewidth}
            \includegraphics[width=\linewidth]{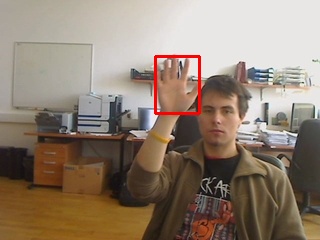}
            \includegraphics[width=\linewidth]{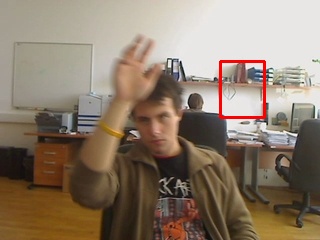}
            %\caption{MAVOT}
        \end{subfigure}
    \end{subfigure}
    \begin{subfigure}{0.98\linewidth}
        \begin{subfigure}{0.32\linewidth}
            \includegraphics[width=\linewidth]{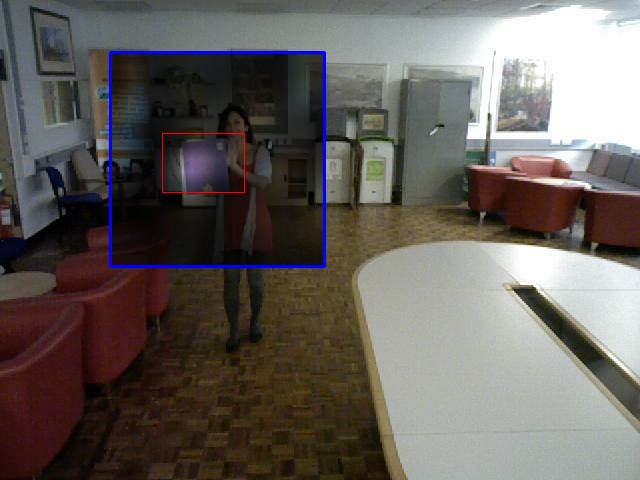}
            \includegraphics[width=\linewidth]{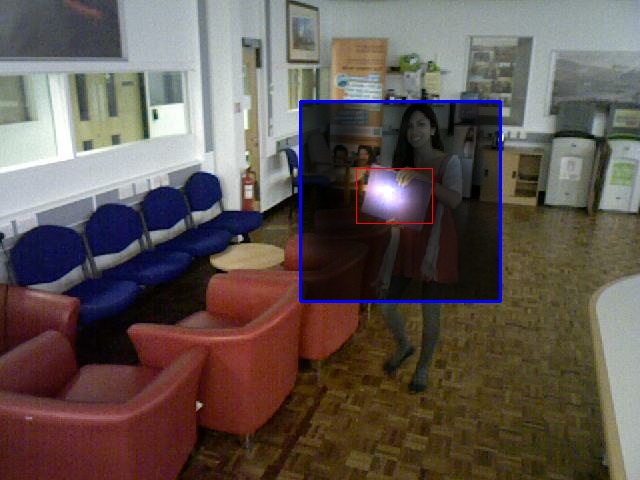}
            \caption{DAWN}
        \end{subfigure}
        \begin{subfigure}{0.32\linewidth}
            \includegraphics[width=\linewidth]{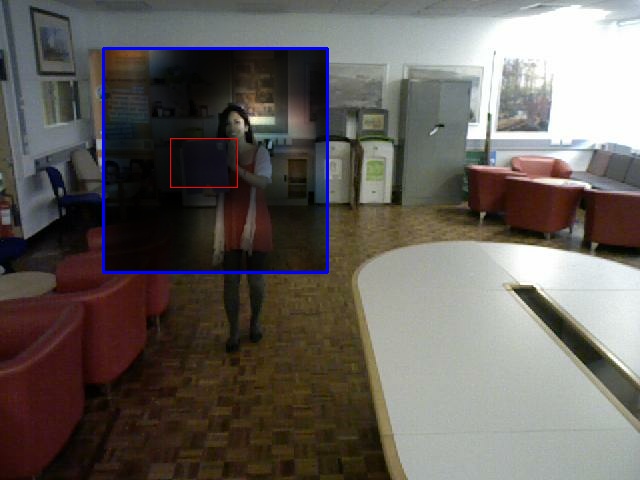}
            \includegraphics[width=\linewidth]{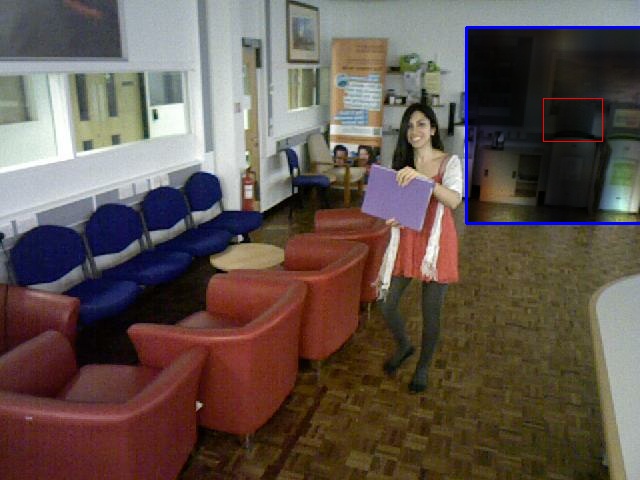}
            \caption{Memtrack}
        \end{subfigure}
        \begin{subfigure}{0.32\linewidth}
            \includegraphics[width=\linewidth]{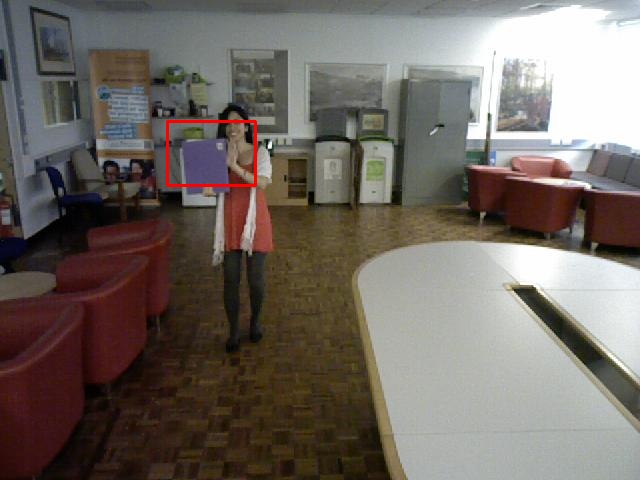}
            \includegraphics[width=\linewidth]{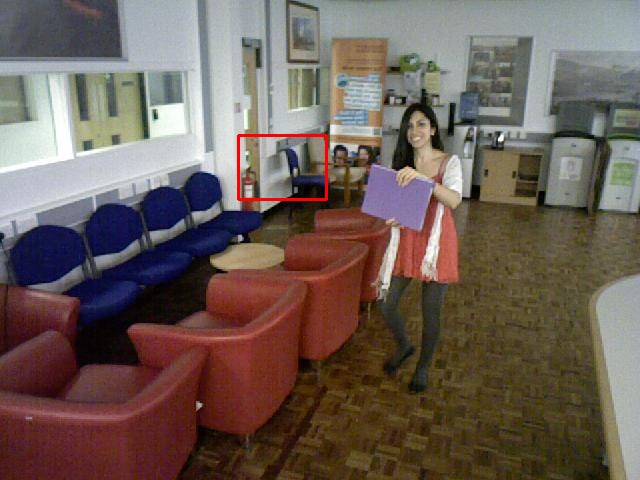}
            \caption{MAVOT}
        \end{subfigure}
    \end{subfigure}
    \caption{Results of two sample frames selected in \textit{hand} and \textit{book} respectively. Time axis is running downward.}
    \label{fig:handnbook}
\vspace{-0.25in}
\end{figure}

Figure~\ref{fig:handnbook} shows the respective attention of DAWN and MemTrack: the blue box is the ROI where the attention is computed, and the inner box (red) is the bounding box produced by the respective tracker. Within a given ROI, brightness indicates the scale of the attention: the brighter the pixel, the higher the attention at that pixel.

Observing the results on the two videos `hand' and `book', when tracking starts, MemTrack's attention is quite random and mostly focused on the background, causing it to lose track of the object in subsequent frames due to a background object with similar features (in video `hand' under severe motion blur both the face and hand are dominated by flesh color), or changes in the object's relative position on the background (in video `book'). On the other hand, DAWN produces a more precise attention, thus giving a stronger focus on the object in the early stage of tracking and making it less vulnerable to losing track in later frames. A stronger focus helps the model to be more robust against similar objects in video `hand', and a precise attention prevents the model from tracking background features.

\begin{figure}[htb]
    \centering
    \begin{subfigure}[b]{0.98\linewidth}
        \includegraphics[width=0.19\linewidth]{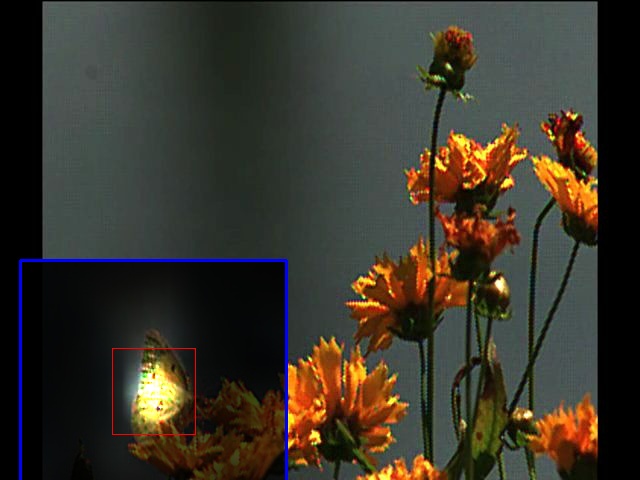}
        \includegraphics[width=0.19\linewidth]{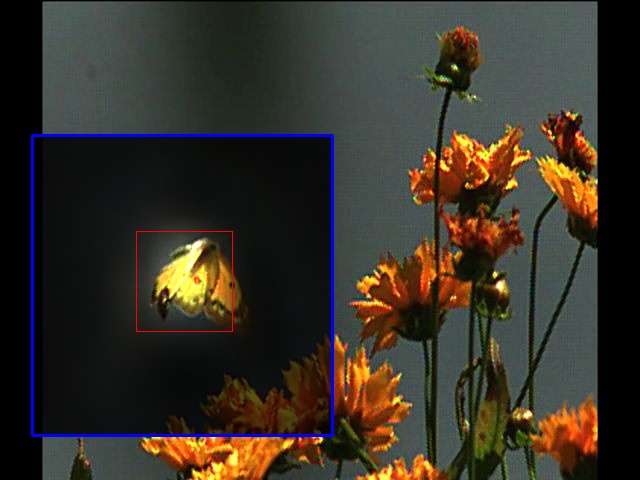}
        \includegraphics[width=0.19\linewidth]{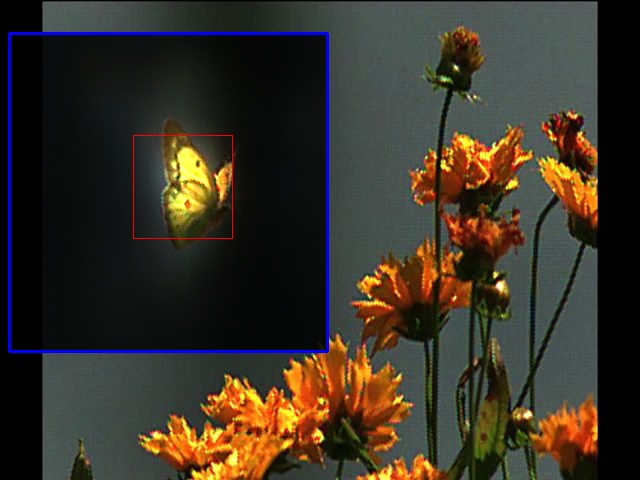}
        \includegraphics[width=0.19\linewidth]{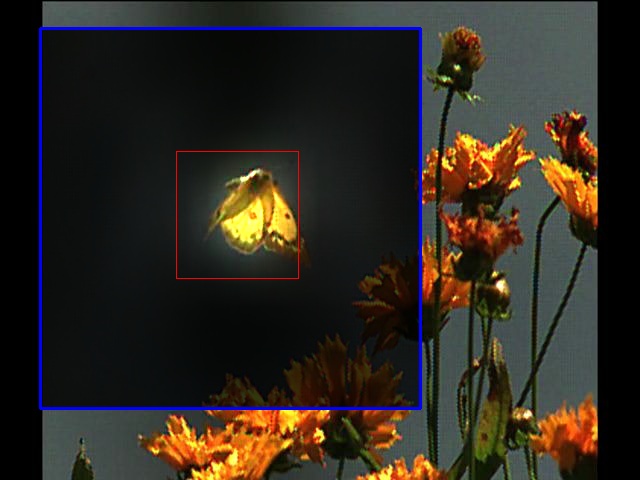}
        \includegraphics[width=0.19\linewidth]{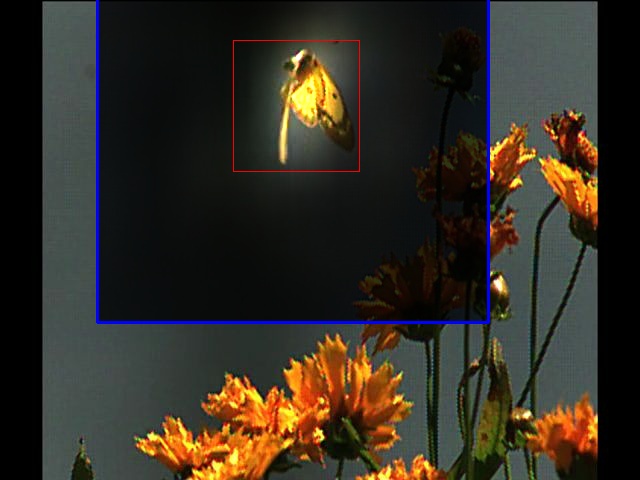}
        \caption{DAWN}
    \end{subfigure}
    \begin{subfigure}[b]{0.98\linewidth}
        \includegraphics[width=0.19\linewidth]{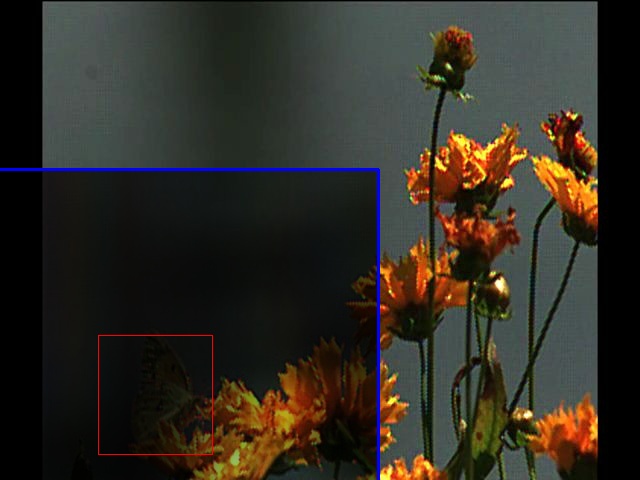}
        \includegraphics[width=0.19\linewidth]{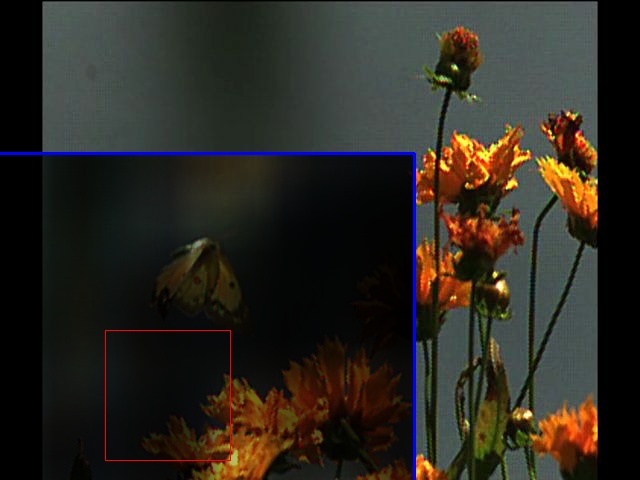}
        \includegraphics[width=0.19\linewidth]{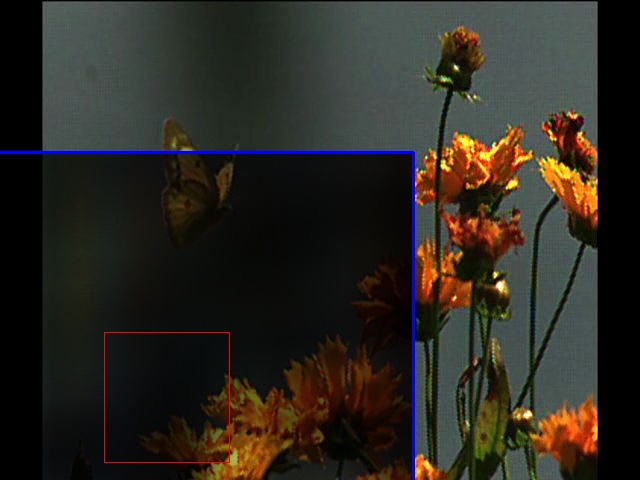}
        \includegraphics[width=0.19\linewidth]{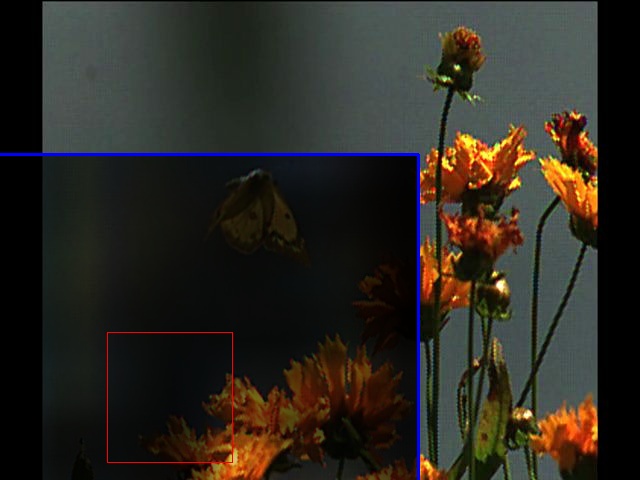}
        \includegraphics[width=0.19\linewidth]{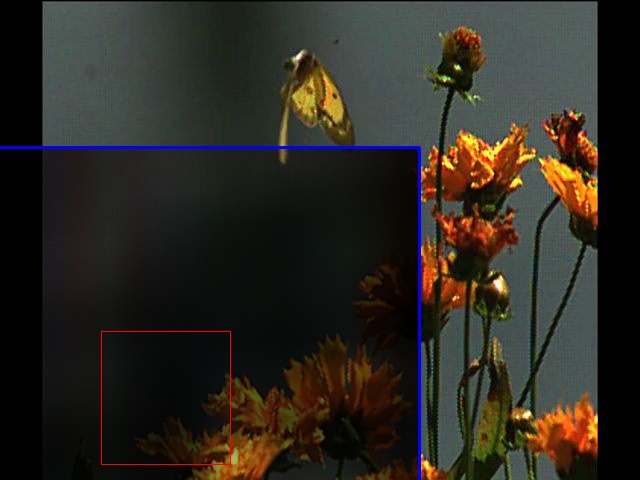}
        \caption{MemTrack}
    \end{subfigure}
    \caption{Attention in \textit{butterfly}.}
\vspace{-0.15in}
    \label{fig:butt}
\end{figure}

In `butterfly' (Figure~\ref{fig:butt}) a false initial attention causes  MemTrack  not moving at all, whereas DAWN with the proper initialization can readily track the object all along.

\begin{figure*}[thb]
    \centering
    \begin{subfigure}[b]{0.90\linewidth}
        \includegraphics[width=0.19\linewidth]{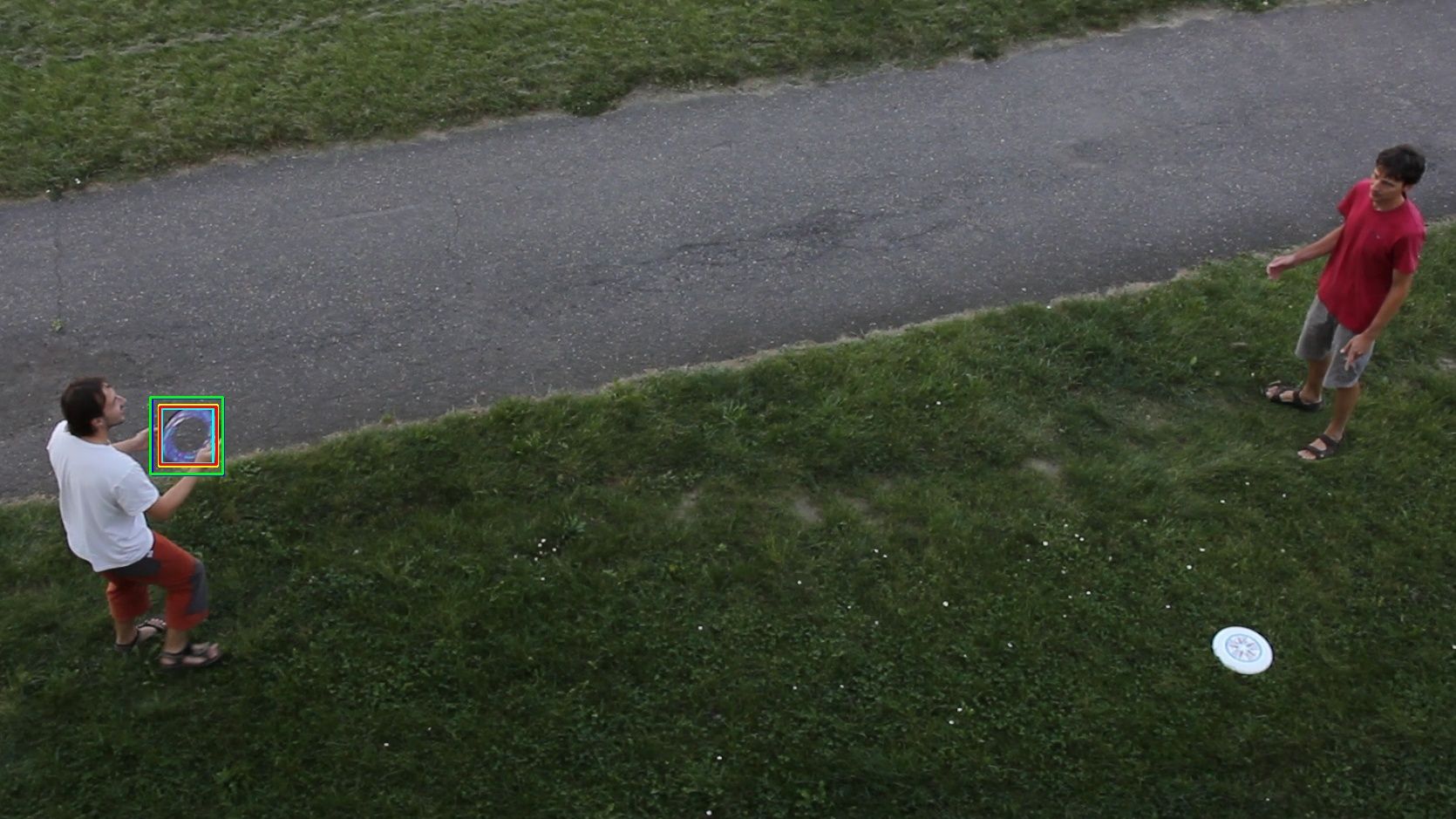}
        \includegraphics[width=0.19\linewidth]{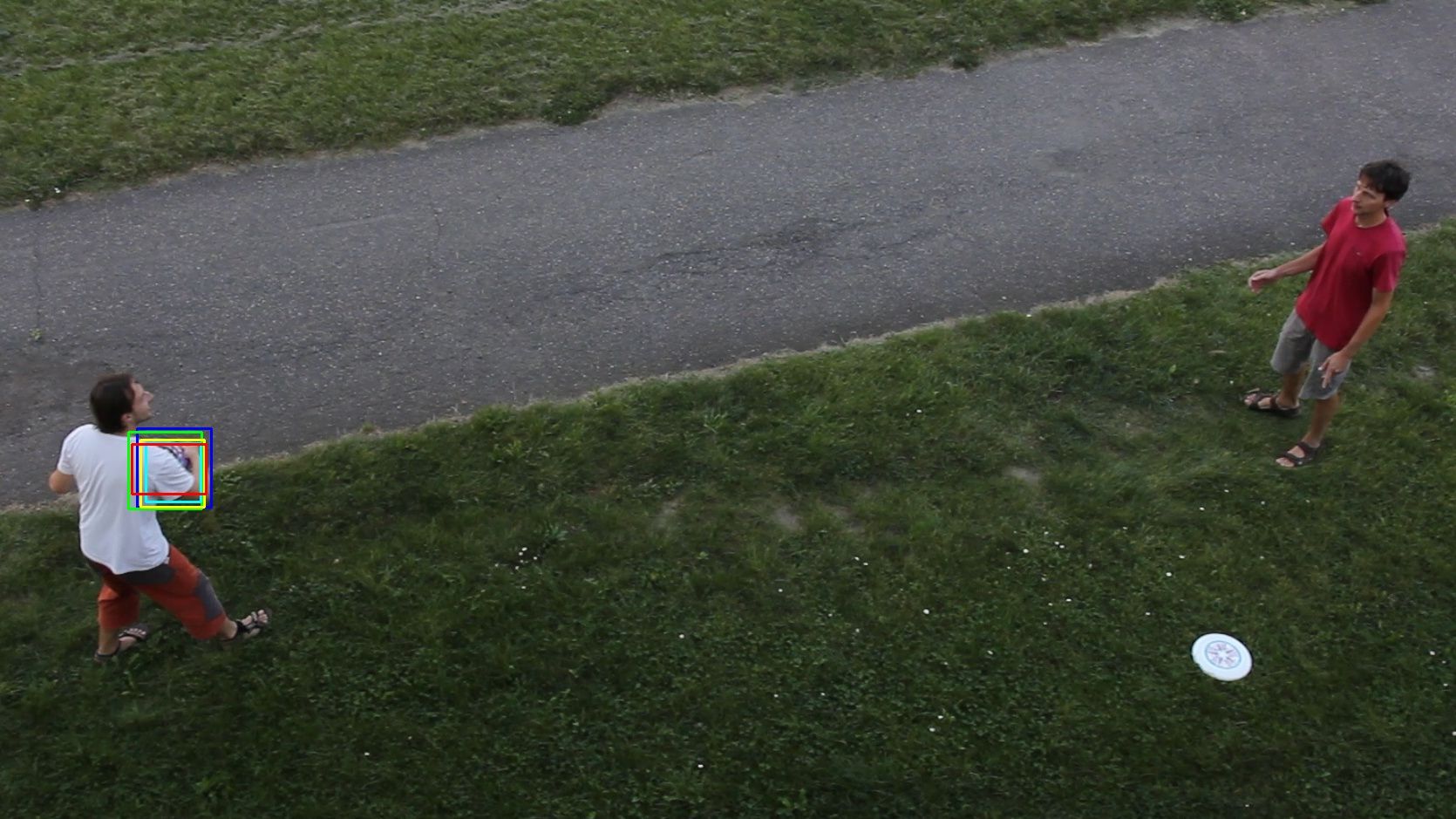}
        \includegraphics[width=0.19\linewidth]{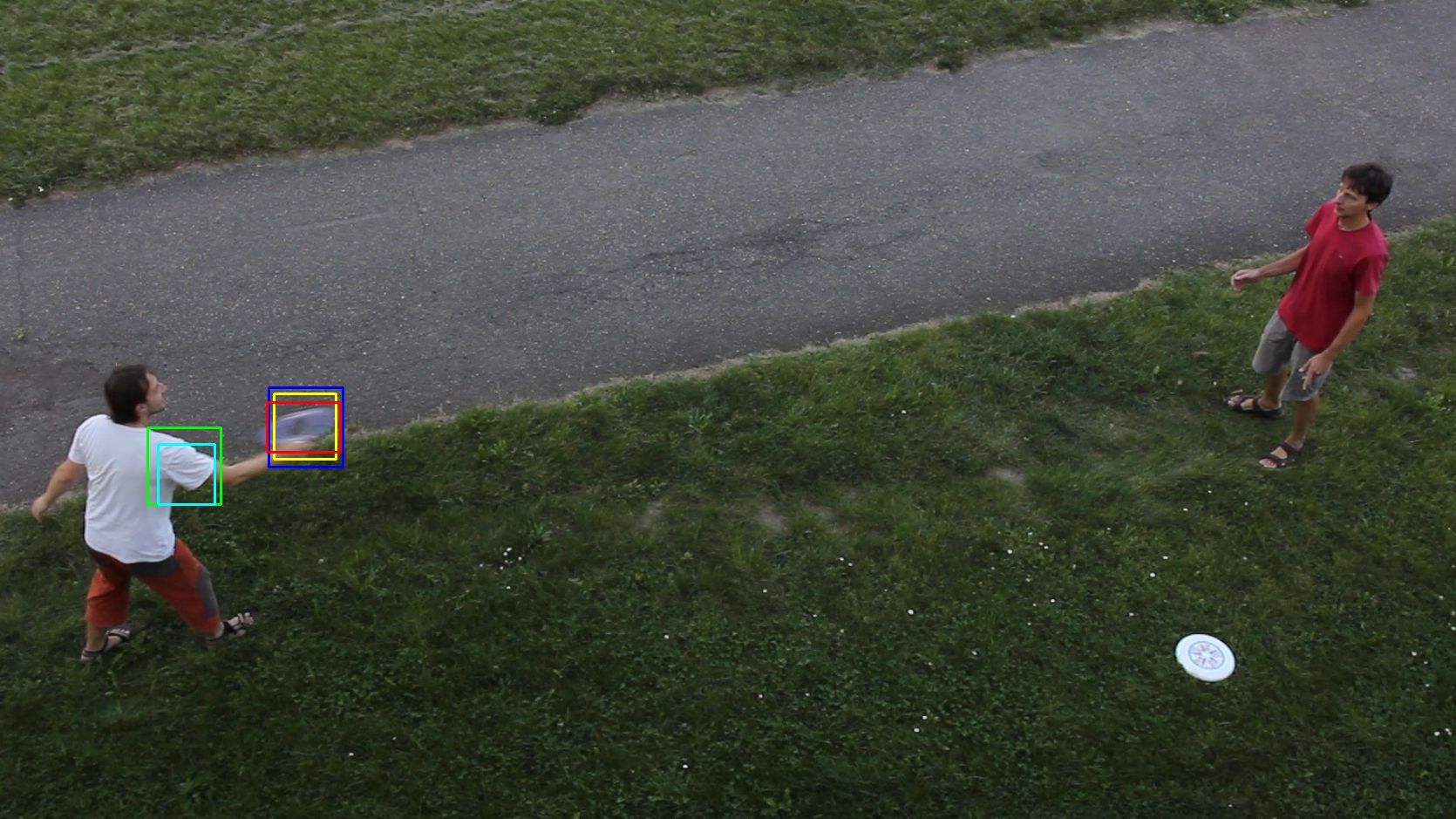}
        \includegraphics[width=0.19\linewidth]{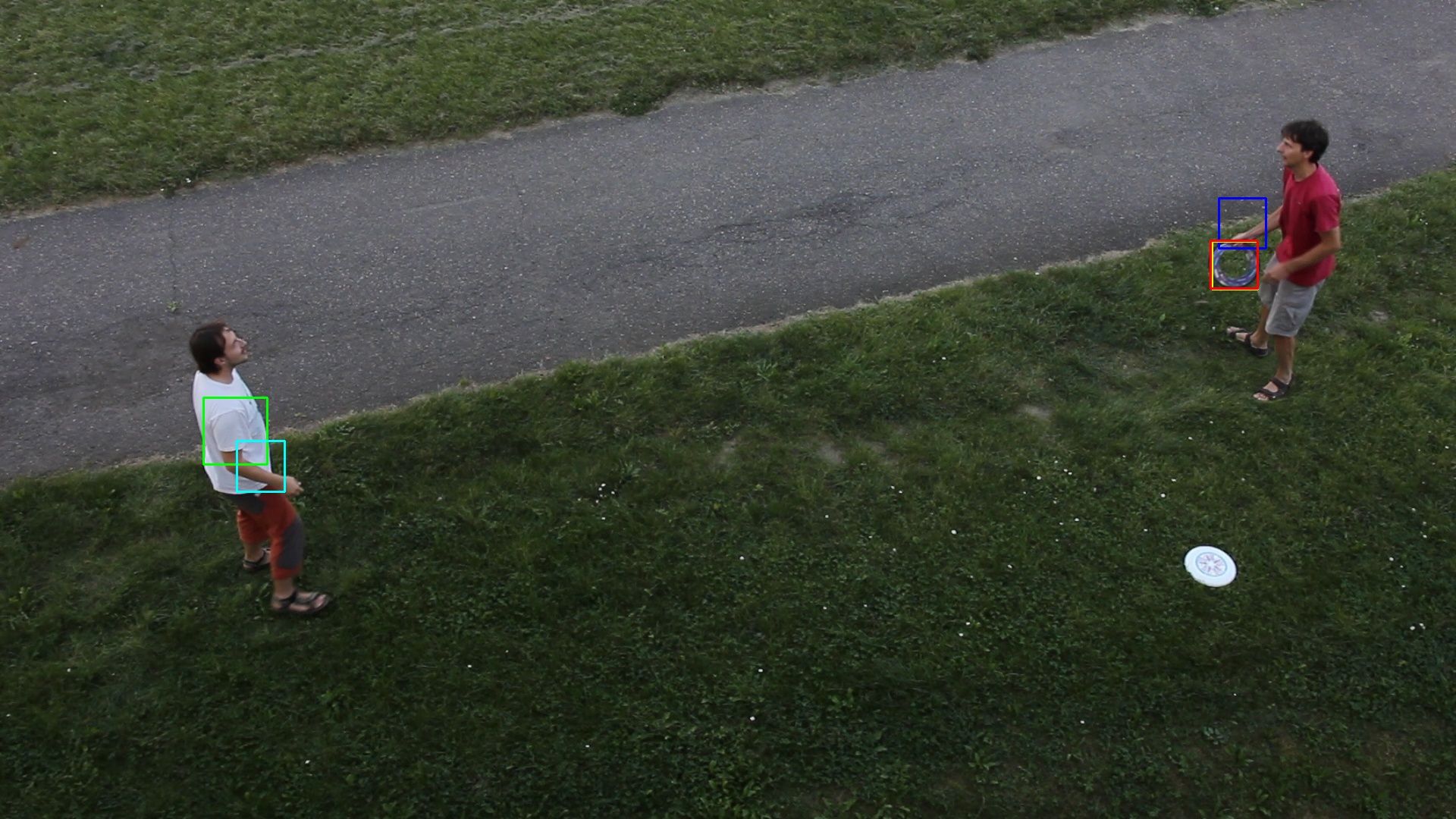}
        \includegraphics[width=0.19\linewidth]{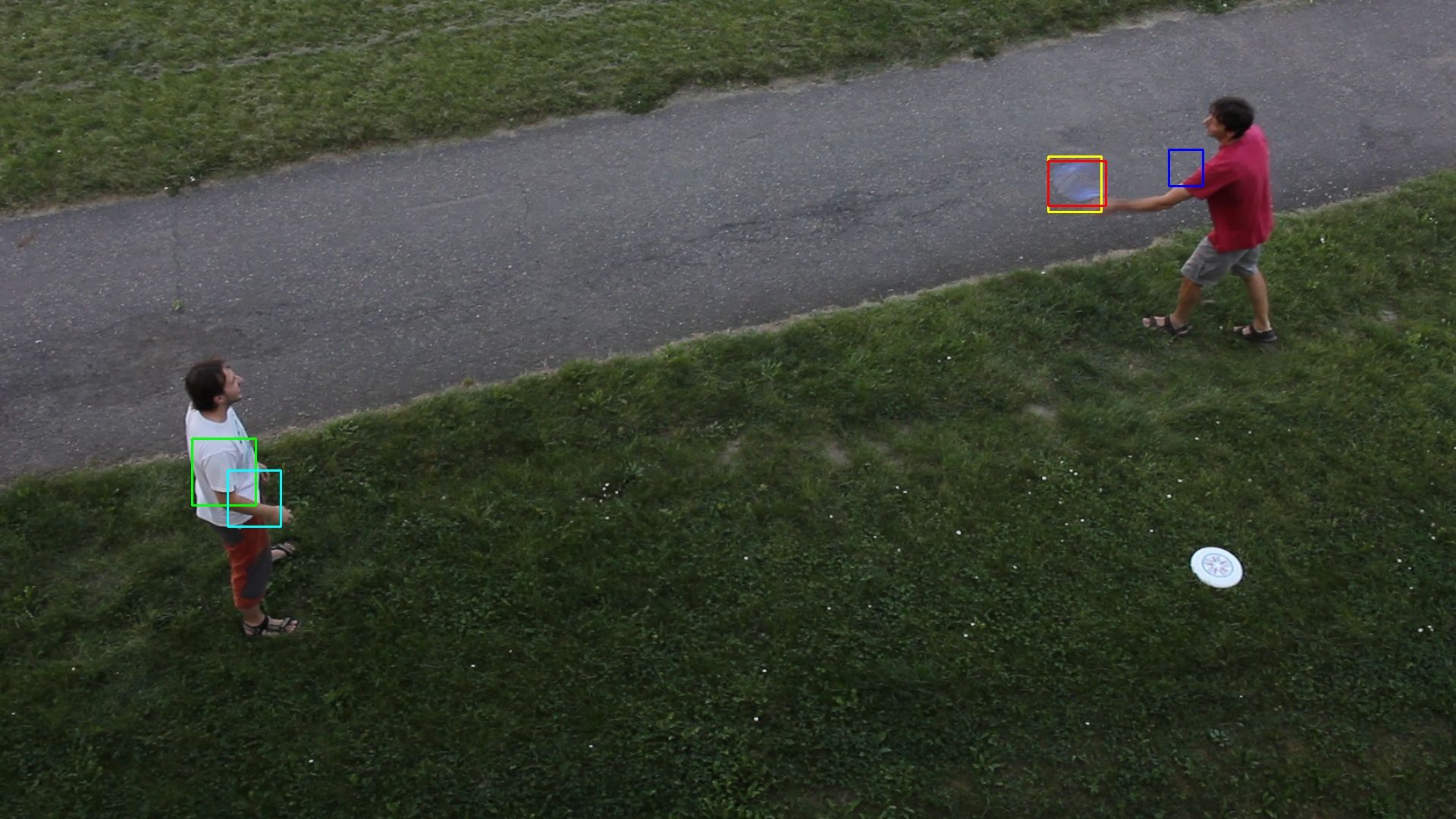}
    \end{subfigure}
    \begin{subfigure}[b]{0.90\linewidth}
        \includegraphics[width=0.19\linewidth]{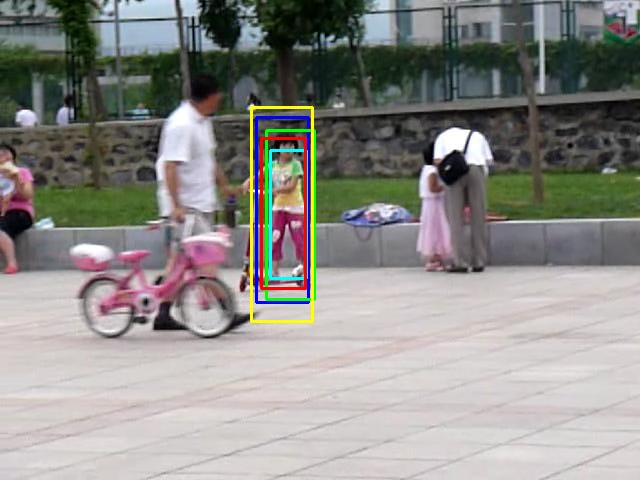}
        \includegraphics[width=0.19\linewidth]{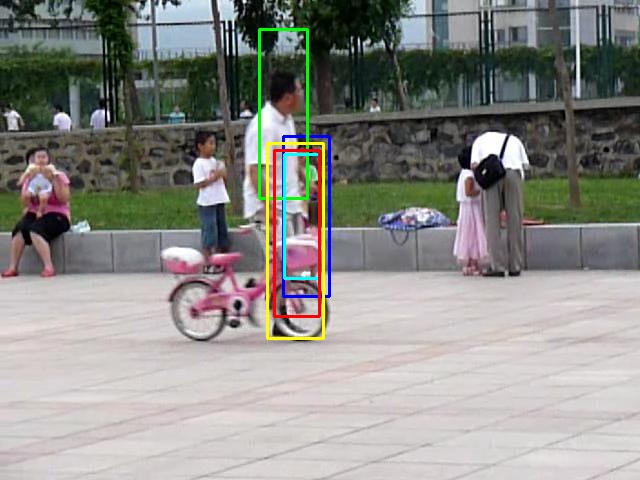}
        \includegraphics[width=0.19\linewidth]{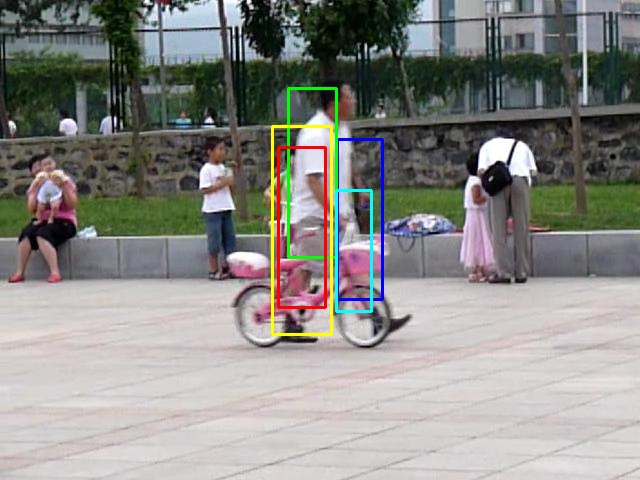}
        \includegraphics[width=0.19\linewidth]{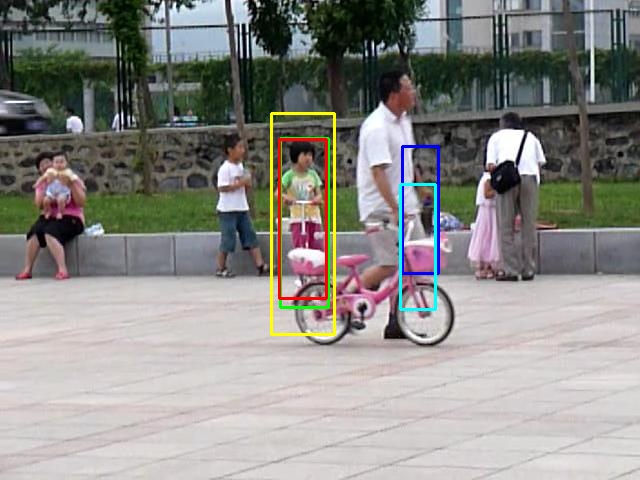}
        \includegraphics[width=0.19\linewidth]{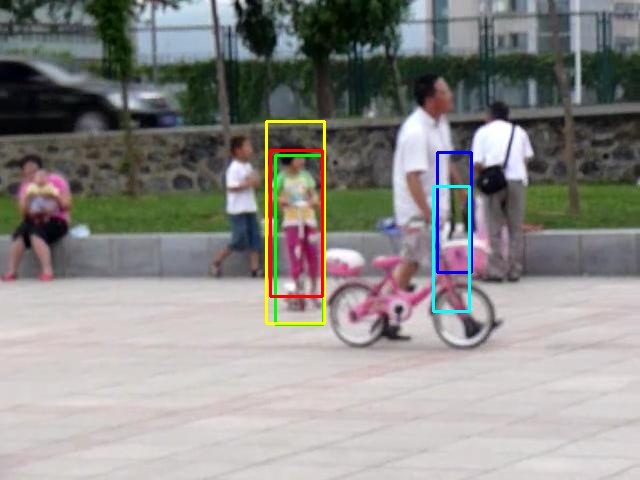}
    \end{subfigure}
    \begin{subfigure}[b]{0.90\linewidth}
        \includegraphics[width=0.19\linewidth]{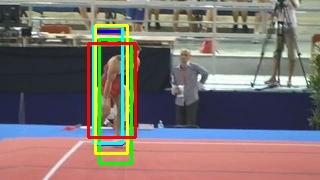}
        \includegraphics[width=0.19\linewidth]{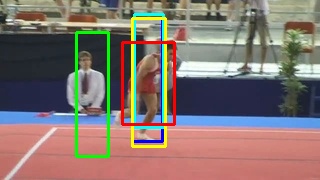}
        \includegraphics[width=0.19\linewidth]{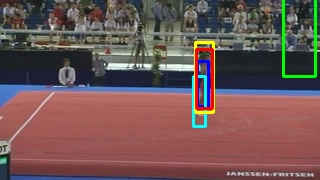}
        \includegraphics[width=0.19\linewidth]{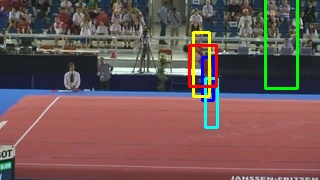}
        \includegraphics[width=0.19\linewidth]{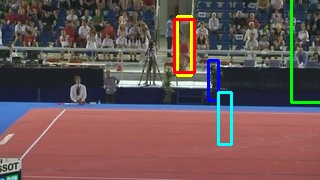}
    \end{subfigure}
    \begin{subfigure}[b]{0.90\linewidth}
        \includegraphics[width=0.19\linewidth]{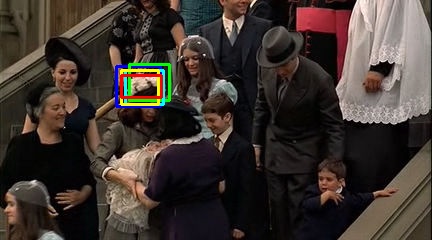}
        \includegraphics[width=0.19\linewidth]{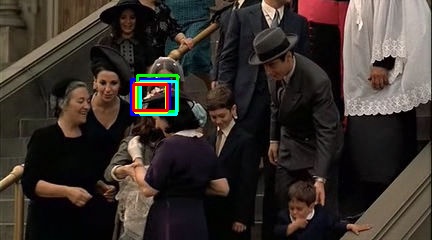}
        \includegraphics[width=0.19\linewidth]{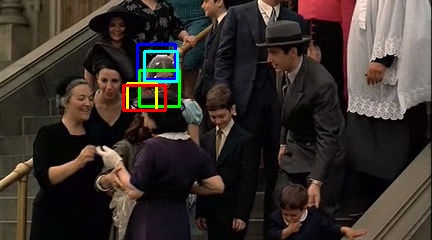}
        \includegraphics[width=0.19\linewidth]{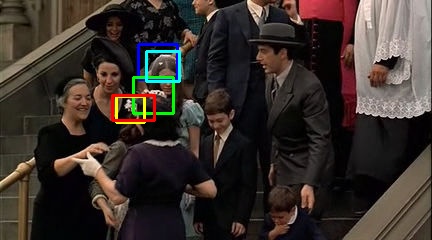}
        \includegraphics[width=0.19\linewidth]{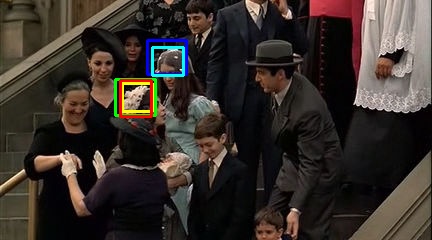}
    \end{subfigure}
    \begin{subfigure}[b]{0.90\linewidth}
        \includegraphics[width=0.19\linewidth]{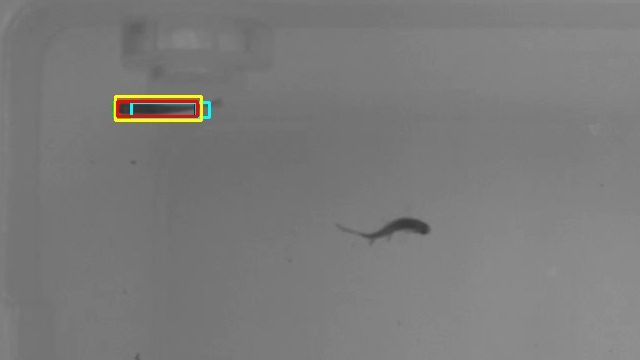}
        \includegraphics[width=0.19\linewidth]{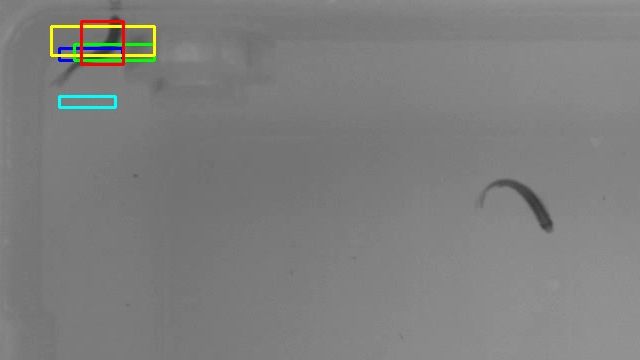}
        \includegraphics[width=0.19\linewidth]{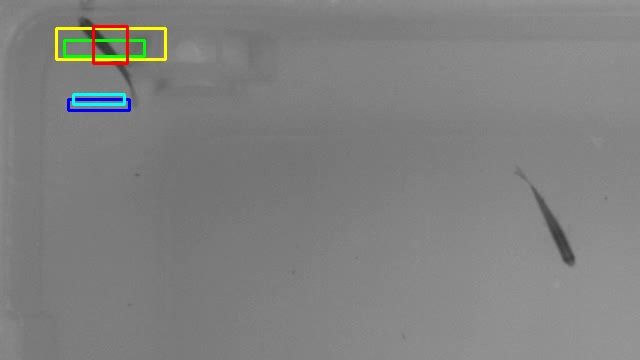}
        \includegraphics[width=0.19\linewidth]{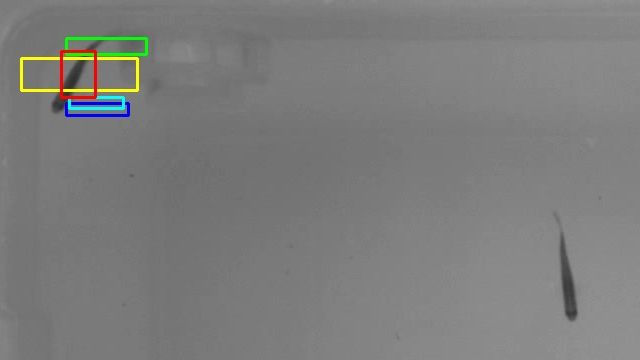}
        \includegraphics[width=0.19\linewidth]{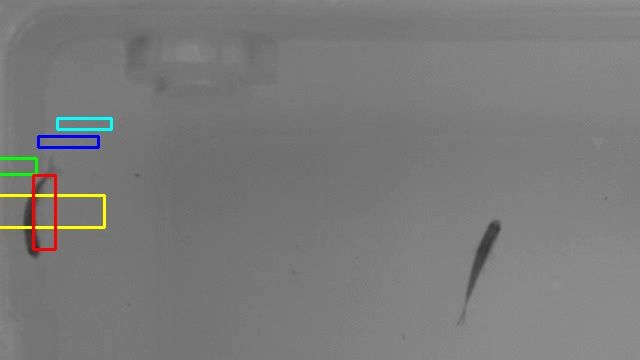}
    \end{subfigure}
    \begin{subfigure}[b]{0.90\linewidth}
        \centering
        \includegraphics[width=0.9\linewidth]{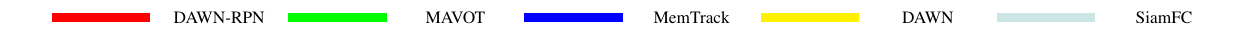}
    \end{subfigure}
    \caption{From top to bottom: \textit{frisbee, girl,  gymnastics1,  godfather, zebrafish1}. }
    \label{fig:mem}
\vspace{-0.2in}
\end{figure*}

\subsection{Memory}
We present a number of examples to show how background memory can effectively address occlusion, confusing background features, multiple object instances and abrupt changes of appearance (Figure~\ref{fig:mem}).

Both partial and total occlusion occur in `frisbee' and `girl'. In both videos, DAWN's background memory module can suppress the occluder (i.e., the guy), allowing the original target to be correctly tracked even after total occlusion. In `godfather', the white veil and white hat distract the trackers without background memory. But DAWN still works well thanks to its effective background memory. Notice that due to abrupt changes of target appearance and similar background features, tracking in `zebrafish1' becomes an almost impossible task for most of the tested trackers, especially those with a fixed aspect ratio of the bounding box. However, with attention and dual memory, DAWN can robustly track the zebrafish, despite the low overlap ratio due to its fixed aspect ratio.

\begin{figure}[!h]
    \centering
    \includegraphics[width=0.14\linewidth]{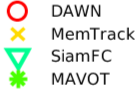}
    \includegraphics[width=0.42\linewidth]{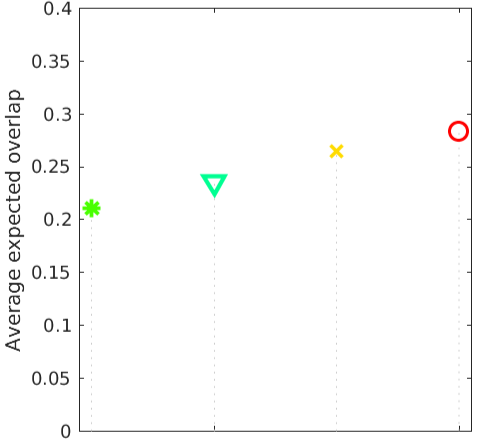}
    \includegraphics[width=0.42\linewidth]{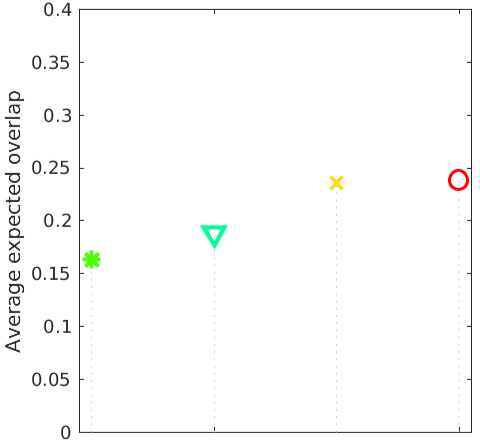}
    \caption{Comparison result of average overlap with ground-truth in VOT2016 (left) and VOT2017 (right).}
    \label{fig:aeo}
    \vspace{-0.20in}
\end{figure}

\begin{table}[h]
  \centering
\begin{adjustbox}{max width=0.8\linewidth}
\begin{tabular}{l|ccc}
\hline
Tracker       & \tabincell{c}{Foreground \\ Memory}  & \tabincell{c}{Background \\ Memory} & Attention \\
\hline
DAWN           & \checkmark & \checkmark & \checkmark \\
MemTrack       & \checkmark &  & \checkmark \\
MAVOT          & \checkmark & \checkmark &  \\
SiamFC         &  &  &  \\
\hline
\end{tabular}
\end{adjustbox}
\caption{Model architecture comparison.}
\label{table:model}
\vspace{-0.1in}
\end{table}

\begin{table}[!t]
  \centering
\begin{adjustbox}{max width=0.8\linewidth}
% \begin{tabular}{l|cccc}
% \hline
%       EAO & SiamRPN  & DaSiam & SA-Siam & DAWN \\
% \hline
% VOT 2016 & 0.34 & 0.41 & 0.29 & 0.28\\
% VOT 2017 & 0.24 & 0.32 & 0.23 & 0.24\\
% \hline
% \end{tabular}
% \caption{STOA Performance}
% \label{table:stoa}
%\end{table}
%\begin{table}[!t]
  \centering
\begin{tabular}{l|cccc}
\hline
      Measures & SiamFC  & MemTrack & DAWN- & DAWN \\
\hline
EAO ($\uparrow$)           & 0.19 & 0.23 & 0.24 & 0.24 \\
Failures ($\downarrow$)      &
34.03 & 30.53 & 27.93 & 26.58 \\
\hline
\end{tabular}
\end{adjustbox}
\caption{Performance in VOT 2017, SiamFC: no memory, MemTrack: target memory, DAWN-: dual memory, DAWN: dual memory with improved attention.}
\label{table:ablation}\vspace{-0.2in}
\end{table}

Table~\ref{table:model} compares the attention and memory modules among memory-based trackers. Figures~\ref{fig:curve} and~\ref{fig:aeo} show our clear improvement over others while still running very fast. Tested in the VOT toolkit, DAWN runs at 15~fps, SiamFC~\cite{SiamFC} at 36 fps,  MemTrack~\cite{memtrack} at 27 fps, and MAVOT~\cite{mavot} at 7 fps. Table~\ref{table:ablation} shows that after adding background memory block and using our new attention scheme the success rates are improved.

%------------------------------------------------------------------------
\vspace{-0.1in}
\section{Conclusion}
 
We have presented a simple and yet effective memory-based approach for unsupervised video object tracking. While strikingly simple in implementation, attention LSTM and dual memory structures enable DAWN to track video objects in unsupervised manner with excellent performance and robustness under challenging scenarios: partial and total occlusion, severe motion blur, abrupt changes in target appearance, multiple object instances, and similar foreground and background features. We have presented extensive quantitative and qualitative experimental comparison: DAWN is ranked third in both VOT2016 and VOT2017 challenges with an excellent success rate among all fast trackers running at fps $>$ 10 and has great performance in unsupervised tracking in both challenges. We further showed using DAWN-RPN that state-of-the-art models can immediately benefit by simply augmenting them with the proposed dual memory and attention LSTM.

 \section*{Acknowledgement}
This research is supported by Tencent and the Research Grant Council of Hong Kong SAR under grant no. 16201818.

{\small
\bibliographystyle{ieee}
\bibliography{egbib}
}

\end{document}